\documentclass{article} % For LaTeX2e
\usepackage{iclr2025_conference,times}

\usepackage{amsmath,amsfonts,bm}

\def\eqref#1{equation~\ref{#1}}
\def\1{\bm{1}}

\DeclareMathAlphabet{\mathsfit}{\encodingdefault}{\sfdefault}{m}{sl}
\SetMathAlphabet{\mathsfit}{bold}{\encodingdefault}{\sfdefault}{bx}{n}

\DeclareMathOperator*{\argmax}{arg\,max}

\usepackage{hyperref}
\usepackage{url}
\usepackage{subcaption}
\usepackage{graphicx}
\usepackage{booktabs}       % professional-quality tables
\usepackage{amsfonts}       % blackboard math symbols
\usepackage[T1]{fontenc}
\usepackage{lmodern}
\usepackage{multirow}
\usepackage{wrapfig}
\usepackage{tikz}
\usetikzlibrary{shapes.geometric, arrows}
\usepackage{amsmath}
\usepackage{algorithm}
\usepackage[noend]{algpseudocode}
\usepackage{appendix}
\usepackage{float}

\tikzstyle{rectgreen} = [rectangle, rounded corners, minimum width=2cm, minimum height=1.5cm,text centered, fill=green!30]
\tikzstyle{rectblue} = [rectangle, rounded corners, minimum width=2cm, minimum height=1.5cm,text centered, fill=blue!30]
\tikzstyle{crclwhite} = [circle, minimum width=1.5cm, text centered, draw=black, fill=white!30]
\tikzstyle{crclorange} = [circle, minimum width=1.5cm, text centered, fill=orange!30]
\tikzstyle{crclyellow} = [circle, minimum width=1.5cm, text centered, fill=yellow!30]
\tikzstyle{dmnd} = [diamond, minimum width=1.5cm, minimum height=1.5cm, text centered, fill=red!30]
\tikzstyle{arrow} = [thick,->,>=stealth]

\newcommand{\MethodName}{Hard View Pretraining}
\newcommand{\MethodAbbr}{HVP}

\title{Beyond Random Augmentations: \\Pretraining with Hard Views}

\author{%
  Fabio Ferreira\thanks{Equal contribution. Correspondence to: ferreira@cs.uni-freiburg.de} \\
  University of Freiburg
  \And 
  Ivo Rapant\footnotemark[1] \\
  University of Freiburg  
  \And 
  J{\"o}rg K.H. Franke \\
  University of Freiburg
  \And
  Frank Hutter \\
  ELLIS Institute T\"ubingen \& \\University of Freiburg\\
}

\iclrfinalcopy % Uncomment for camera-ready version, but NOT for submission.
\begin{document}

\maketitle

\begin{abstract}
Self-Supervised Learning (SSL) methods typically rely on random image augmentations, or \emph{views}, to make models invariant to different transformations. We hypothesize that the efficacy of pretraining pipelines based on conventional random view sampling can be enhanced by explicitly selecting views that benefit the learning progress. A simple yet effective approach is to select \emph{hard views} that yield a higher loss. In this paper, we propose \emph{\MethodName{} (\MethodAbbr{})}, a learning-free strategy that extends random view generation by exposing models to more challenging samples during SSL pretraining. \MethodAbbr{} encompasses the following iterative steps: 1) randomly sample multiple views and forward each view through the pretrained model, 2) create pairs of two views and compute their loss, 3) adversarially select the pair yielding the highest loss according to the current model state, and 4) perform a backward pass with the selected pair. In contrast to existing hard view literature, we are the first to demonstrate hard view pretraining's effectiveness at scale, particularly training on the full ImageNet-1k dataset, and evaluating across multiple SSL methods, ConvNets, and ViTs. As a result, \MethodAbbr{} sets a new state-of-the-art on DINO ViT-B/16, reaching 78.8\% linear evaluation accuracy (a 0.6\% improvement) and consistent gains of 1\% for both 100 and 300 epoch pretraining, with similar improvements across transfer tasks in DINO, SimSiam, iBOT, and SimCLR.
\end{abstract}

\section{Introduction}
% General sentence about two categories in SSL
% Various approaches to learning effective and generalizable visual representations in Self-Supervised Learning (SSL) exist. One way to categorize many SSL methods is to distinguish generative and discriminative approaches \cite{chen-icml20b}. While generative methods aim at generating image input, discriminative methods, with contrastive learning \cite{hadsell-cvpr06a, he-cvpr20a} being the most known approach, aim at learning a latent representation in which similar image views are located closely and dissimilar ones distantly. 

Learning effective and generalizable visual representations in Self-Supervised Learning (SSL) has been approached in various ways. Many SSL methods can be broadly categorized into generative and discriminative approaches \citep{chen-icml20b}. Generative methods focus on generating image inputs, while discriminative methods, particularly contrastive learning \citep{hadsell-cvpr06a, he-cvpr20a}, aim at learning latent representations in which similar image views are located closely, and dissimilar ones distantly.

% Explain crops and how approaches almost all rely on these
% Such views are generated by applying a sequence of (randomly sampled) image transformations and are usually composed of geometric (cropping, rotation, etc.) and appearance (color distortion, blurring, etc.) transformations. A body of literature \cite{chen-icml20b, wu-arxiv20a,purushwalkam-neurips20a, wagner-icml22a, tian-neurips20a} has illuminated the effects of image views on representation learning and identified \emph{random resized crop} (RRC) transformation, which randomly crops the image and resizes it back to a fixed size, as well as color distortion as one of the essential augmentations for effective discriminative learning. However, despite this finding and to our best knowledge, little research has gone into identifying more effective ways for generating views to improve performance. In this paper, we seek to extend the random view generation to expose the model to harder samples during discriminative pretraining by selecting views that yield a high loss according to the current model state.
Such views are generated by applying a sequence of (randomly sampled) image transformations and are usually composed of geometric (cropping, rotation, etc.) and appearance (color distortion, blurring, etc.) transformations. Prior work \citep{chen-icml20b, wu-arxiv20a, purushwalkam-neurips20a, wagner-icml22a, tian-neurips20a} has identified \emph{random resized crop} (RRC), which randomly crops the image and resizes it back to a fixed size, as well as color distortion as critical transformations for effective representation learning. However, despite this finding and to our best knowledge, little research has gone into identifying more effective ways for generating views to improve performance.

\begin{figure}
    \centering
    \begin{subfigure}[b]{0.58\linewidth}
        \centering
        \includegraphics[width=\linewidth]{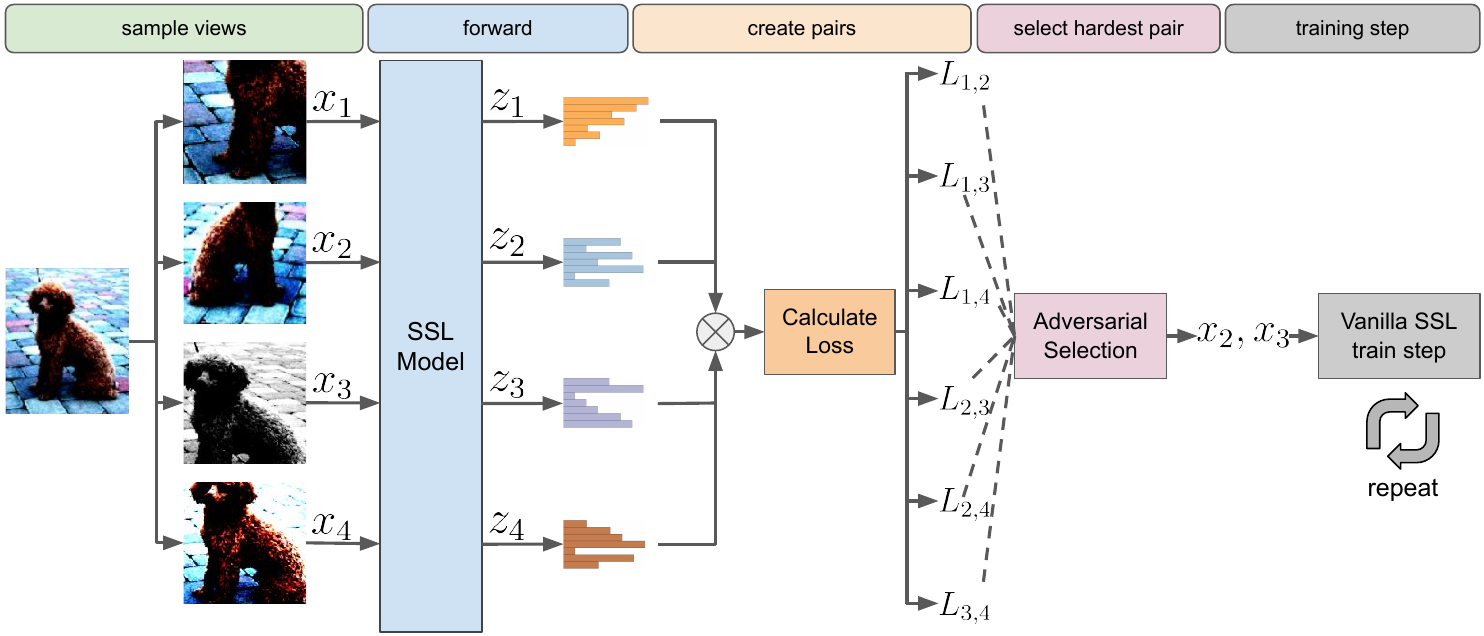}
        \caption{}
        \label{fig:method_overview}
    \end{subfigure}
    \hfill
    \begin{subfigure}[b]{0.4\linewidth}
        \centering
        \includegraphics[width=\linewidth]{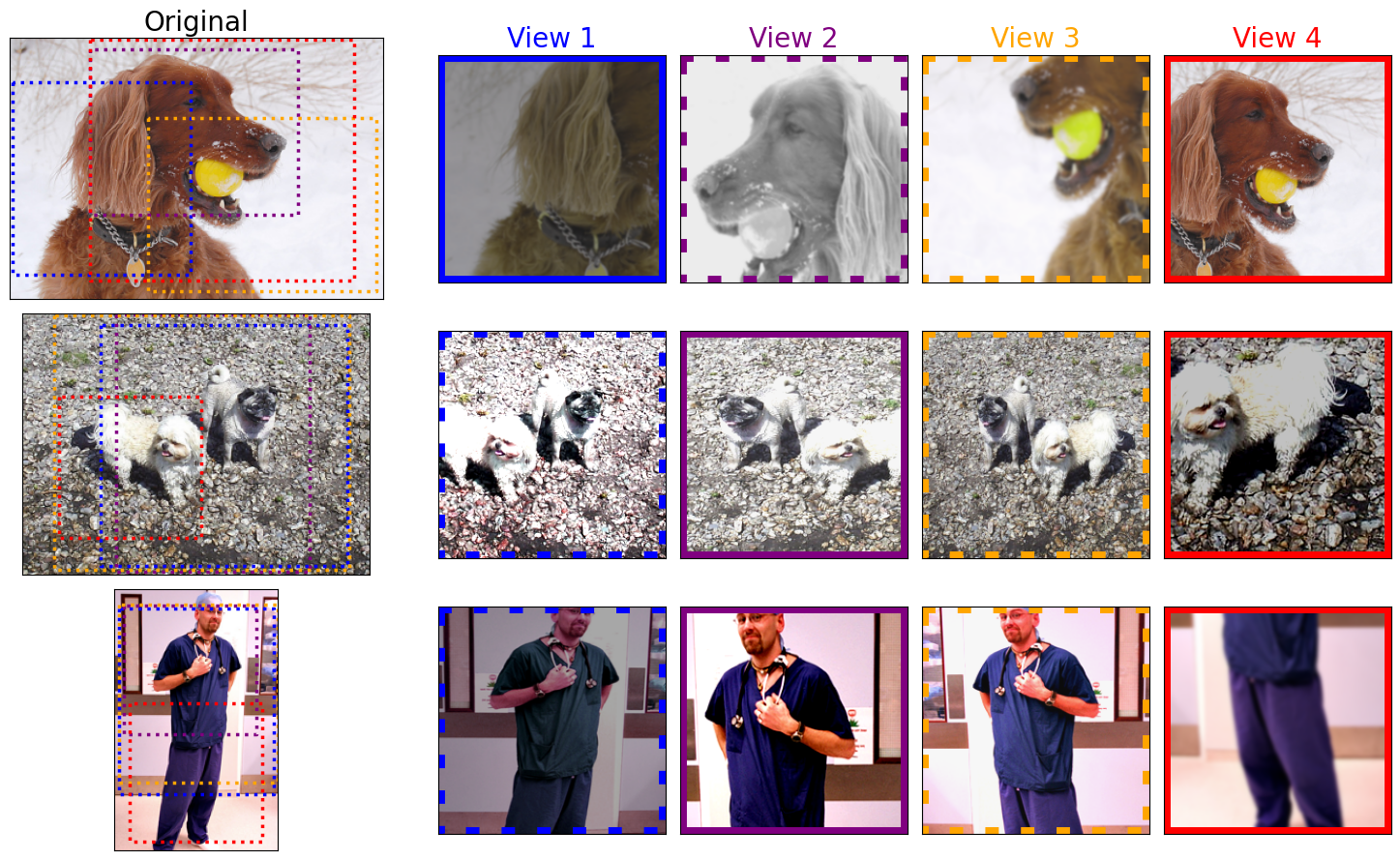}
        \caption{}
        \label{fig:example_views}
    \end{subfigure}
    \caption{\textbf{(a)} \MethodAbbr{} first samples $N$ views, pairs them, and adversarially selects the hardest pair, i.e., the one with the worst loss according to the current model state. \textbf{(b)} Examples (left) and sampled views (right) after transformations. Hard pairs selected by \MethodAbbr{} are shown with a solid frame.}
    \label{fig:combined_images}
\end{figure}

% Related SSL approaches exist that aim at controlling view hardness through adversarial \cite{shi-icml22a, tamkin-iclr21a} or cooperative \cite{hou-arxiv23a} models. A work that takes a similar perspective to ours in leveraging the information content of views for SSL is \cite{tian-neurips20a}. Motivated by mutual information theory, they derive an objective to adversarially learn view generators for contrastive learning. While such approaches offer important insights into view optimality, they typically require learning additional models before the actual SSL training or involve non-trivial changes to the pipeline, making them less easily integrated into existing SSL methods. Moreover, these methods are unlikely to find their way into current state-of-the-art models, as training such models is already resource-intensive, and simpler methods have proven more practical and effective.
Existing SSL approaches that attempt to control the hardness of views include adversarial \citep{shi-icml22a, tamkin-iclr21a} or cooperative \citep{hou-arxiv23a} techniques. For instance, \cite{tian-neurips20a} use mutual information theory to adversarially learn view generators. These methods offer valuable insights into optimizing views but often introduce additional complexity, such as requiring additional models or significant changes to the SSL pipeline, limiting their practicality in state-of-the-art models where resource constraints are already a concern. Similar to what we propose, \cite{kocyigit-wacv23} introduce a hard view sampling strategy but targeting the acceleration of pretraining. Moreover, their approach requires tuning hyperparameters like learning rate and augmentation magnitude, which can introduce confounding variables. Furthermore, neither \cite{kocyigit-wacv23} nor \cite{tian-neurips20a} validate their methods on larger datasets such as ImageNet-1k \citep{deng-cvpr09a}, or across larger model architectures, limiting their scalability and applicability.

Building on these observations, we propose \emph{\MethodName{} (\MethodAbbr{})}, a fully learning-free, easy-to-integrate approach designed to improve standard pretraining methods without the need for additional model training or complex modifications. Our method leverages the current model state to select challenging samples during pretraining by adversarially sampling pairs of views and selecting the pair that yields the highest loss according to the model`s current state (see Fig. \ref{fig:method_overview}). Unlike previous approaches, \MethodAbbr{} requires no hyperparameter tuning and demonstrates scalability to large datasets like ImageNet-1k, offering an efficient and practical solution for improving SSL pretraining. To the best of our knowledge, we are the first to demonstrate the effectiveness of a hard view sampling strategy at scale, particularly on modern architectures like Vision Transformers (ViTs) and training on the full ImageNet dataset. Our approach not only integrates seamlessly with recent state-of-the-art SSL methods but also showcases consistent improvements across both convolutional architectures and ViTs, validating its robustness and scalability.

% We propose a fully learning-free, easy-to-integrate approach that improves over standard pretraining methods. Our method leverages the current model state and operates on a per-sample basis during pretraining. Specifically, we sample a few views, forward each pair through the model, and identify the pair yielding the highest loss for the backward pass, repeating this process throughout training (see Fig. \ref{fig:method_overview}). While being agnostic to the loss, this strategy exposes the pretrained model to samples that the model at any given point in the training trajectory finds challenging. As our results show, sampling only four views is sufficient to achieve noticeable improvements.

% A related approach is \cite{kocyigit-wacv23}, which also employs a hard view strategy. However, their approach targets speeding up pretraining and requires tuning hyperparameters like learning rate and augmentation magnitude, which can introduce confounding factors. More critically, neither \cite{kocyigit-wacv23} nor \cite{tian-neurips20a} demonstrate performance on larger datasets like ImageNet-1k \cite{deng-cvpr09a} or across larger model architectures. In contrast, we are the first to show that the approach of hard view pretraining works on large-scale models and datasets without confounding factors such as hyperparameter tuning. Our method requires no hyperparameter adjustments and is validated at scale, including longer pretraining schedules, the full ImageNet dataset, and larger models, showcasing its practical effectiveness and scalability.

Overall, our contributions can be summarized as follows:
\begin{itemize}
    \item We propose \emph{\MethodName{} (\MethodAbbr{})}, an easy-to-use method complementing SSL by extending the common random view generation to automatically expose the model to harder samples during pretraining. \MethodAbbr{} simply requires the ability to compute sample-wise losses;
    \item We demonstrate the effectiveness and compatibility of our approach using ImageNet-1k pretraining across four popular SSL methods that cover a diverse range of discriminative objectives such as SimSiam \citep{chen-cvpr21b}, DINO \citep{caron-iccv21a}, iBOT \citep{zhou-icml22a}, and SimCLR \citep{chen-icml20b};
    \item \MethodAbbr{} achieves a new state-of-the-art result on DINO ViT-B/16, improving over the officially reported baseline of 78.2\% linear evaluation accuracy by reaching 78.8\% (400 epochs). \MethodAbbr{} also consistently improves all other baselines by an average of 1\% in linear evaluation on ImageNet across 100 and 300 epoch-pretraining runs;
    \item We show similar improvements on a diverse set of transfer tasks, including finetuning, object detection, and segmentation, and present insights into the underlying mechanisms and robustness of \MethodAbbr{}.
\end{itemize}

We make our PyTorch \cite{pytorch-neurips19a} code, models, and all used hyperparameters publicly available under \url{https://github.com/automl/hvp}.
\section{Related Work}
\subsection{Discriminative Self-Supervised Learning} % could also be named "Augmentation-based SSL"
The core idea behind the discriminative learning framework \citep{chen-icml20b} is to learn image representations by contrasting positive pairs (two views of the same image) against negative pairs (two views of different images) \citep{hadsell-cvpr06a}. To work well in practice and to prevent model collapse, contrastive learning methods often require a large number of negative samples \citep{wu-cvpr18a, vandenoord-arxiv18a, chen-icml20b, he-cvpr20a, tian-eccv20b, chen-arxiv2020a} stored in memory banks \citep{wu-cvpr18a, he-cvpr20a} or, for instance, in the case of SimCLR, implicitly in large batches \citep{chen-icml20b}. Non-contrastive approaches, such as BYOL \citep{grill-neurips20a}, SimSiam \citep{chen-cvpr21b}, DINO \citep{caron-iccv21a} and others \citep{zbontar-icml21a, caron-neurips20a, ermolov-icml21a}, can only use positive pairs without causing model collapse but rely on other techniques, such as Siamese architectures, whitening of embeddings, clustering, maximizing the entropy of the embeddings, momentum encoders, and more. 

\subsection{Optimizing for Hard Views in SSL}
Due to its performance-improving benefits, the realm of learning task-specific augmentation policies based on data has seen quick development \citep{cubuk-cvpr19a, ho-icml19a, lin-iccv19a, zhang-iclr20a, hataya-eccv20a, hou-arxiv23a, muller-iccv21a}. However, these approaches do not include the random resize crop operation in their search spaces, limiting the control of view hardness. Similar to us, \cite{kocyigit-wacv23} uses the current model state for selecting hard views. However, their approach requires controlling learning hyperparameters, while mostly training on a smaller version of ImageNet and ResNets \citep{he-cvpr16a} only. We offer a more complete analysis of hard view pretraining, demonstrating the benefits without potential confounding factors such as hyperparameter adjustments. We also focus on performance rather than pretraining speedups and employ higher budgets (longer pretraining, larger batch sizes) on the full ImageNet dataset and both ResNets and ViTs \citep{dosovitskiy-iclr21a}). Other works utilize additional networks for view generation, such as \cite{tamkin-iclr21a, shi-icml22a, tian-neurips20a} (adversarial view generators), \cite{peng-cvpr22a} (localization network for semantic awareness), \cite{li-eccv24a} (pretrained generative models to enhance augmentation quality), and \cite{han2023constructive} (leveraging a pretrained GAN in a SimCLR-only setting). However, unlike \MethodAbbr{}, all these methods add non-trivial complexity to the training pipeline by requiring learning auxiliary and adversarial components or are often limited in their applicability across different SSL frameworks (e.g., by requiring negative view pairs).
\section{Method}
\subsection{Self-supervised Learning Framework}
In this section, we introduce our approach, which is also depicted in Algorithm \ref{alg}. Many different self-supervised discriminative learning \citep{he-cvpr20a} objectives exist, each characterized by variations stemming from design choices, such as by the use of positive and negative samples or asymmetry in the encoder-projector network structure. For simplicity of exposure, we will introduce our approach based on the SimSiam objective \citep{chen-cvpr21b}, but we do note that our method can be used with any other discriminative SSL objective that allows the computation of sample-wise losses. 

SimSiam works as follows. Assume a given set of images $\mathcal{D}$, an image augmentation distribution $\mathcal{T}$, a minibatch of $M$ images $\mathbf{x} =\left\{x_i\right\}_{i=1}^M$ sampled uniformly from $\mathcal{D}$, and two sets of randomly sampled image augmentations $A$ and $B$ sampled from $\mathcal{T}$. SimSiam applies $A$ and $B$ to each image in $\mathbf{x}$ resulting in $\mathbf{x}^A$ and $\mathbf{x}^B$. Both augmented sets of views are subsequently projected into an embedding space with $\mathbf{z}^A=g_{\theta}(f_{\theta}(\mathbf{x}^A))$ and $\mathbf{h}^B=f_{\theta}(\mathbf{x}^B)$ where $f_{\theta}$ represents an encoder (or backbone) and $g_{\theta}$ a projector network. SimSiam then minimizes the following objective:
\begin{equation}
    \label{eq:simsiam}
    \mathcal{L}(\theta) = \frac{1}{2} \left( D(\mathbf{z}^A, \mathbf{h}^B) + D(\mathbf{z}^B, \mathbf{h}^A) \right)    
\end{equation}
where \emph{D} denotes the negative cosine similarity function. Intuitively, when optimizing $\theta$, the embeddings of the two augmented views are attracted to each other. 

\subsection{Pretraining with Hard Views}
We now formalize how we expose the model to more challenging views during pretraining. In a nutshell, \MethodName{} extends the random view generation by sampling adversarially harder views during pretraining. Instead of having two sets of augmentations $A$ and $B$, we now sample $N$ sets of augmentations, denoted as $\mathcal{A} = \{A_1, A_2, \ldots, A_N\}$. Each set $A_i$ is sampled from the image augmentation distribution $\mathcal{T}$, and applied to each image in $\mathbf{x}$, resulting in $N$ augmented sets of views $\mathbf{x}^{A_1}, \mathbf{x}^{A_2}, \ldots, \mathbf{x}^{A_N}$. Similarly, we obtain $N$ sets of embeddings \(\mathbf{z}^{A_1}, \mathbf{z}^{A_2}, \ldots, \mathbf{z}^{A_N}\) and predictions \(\mathbf{h}^{A_1}, \mathbf{h}^{A_2}, \ldots, \mathbf{h}^{A_N}\) through the encoder and projector networks. We then define a new objective function that seeks to find the pair $(x_{i}^{A_{k}}, x_i^{A_{l}})$ of a given image $x_i$ that yields the highest loss:

\begin{equation}
    \label{eq:acl}
    \begin{aligned}
        (x_i^{A_k}, x_i^{A_l}) &= \argmax_{k, l; k \neq l} \mathcal{L}(\theta)_{i, k,l} \\
        &= \argmax_{k, l; k \neq l} \frac{1}{2} \left( D(z_i^{A_k}, h_i^{A_l}) + D(z_i^{A_l}, h_i^{A_k}) \right),
    \end{aligned}
\end{equation}

%\begin{equation}
%    (x^{A_{k^*}}, x^{A_{l^*}}) = \arg\max_{k, l; k \neq l} \mathcal{L}(\theta)_{*, kl}
%\end{equation}

%\begin{equation}
%    (k^*, l^*) = \arg\max_{k, l: k \neq l} \mathcal{L}(\theta)_{kl} \text{nicht mehr nötig}
%\end{equation}

\noindent{}where $\mathcal{L}(\theta)_{i, k, l}$ simply denotes a sample-wise variant of Eq. \ref{eq:simsiam}.
%where the embeddings are superscripted by $A_{k}$, $A_{l}$ with \(k, l \in \{1,2, \ldots, N\}\) and subscripted by $i$. 

\begin{wrapfigure}[16]{r}{0.58\linewidth}
\vspace{-0.7cm}
\begin{minipage}{\linewidth}
\begin{algorithm}[H]
\begin{algorithmic}[1]
\caption{Pretraining with Hard Views}
%\footnotesize
%\setlength{\itemindent}{0.5em}
\State \textbf{Input:} Number of views $N \geq 2$, batch size $M$, 
\State augmentation distribution \(\mathcal{T}\), model \(f\) %encoder \(f_{\theta}\), projector \(g_{\theta}\)
\For{each $x_i$ in the sampled batch \(\left\{x_i\right\}_{i=0}^M\)}
    \State Sample $N$ augmentations: $A=\left\{t_n \sim \mathcal{T} \right\}_{n=0}^N$
    \State Create augmented views: $\textbf{x}_i^{A}= \left\{t_n(x_i)\right\}_{n=0}^N$ 
    \State Forward all views through $f$ %and $g_{\theta}$ to get $z$ and $h$
    \State Create all $\binom{N}{2}$ view pairs $\textbf{x}_i^{A_k} \times \textbf{x}_i^{A_l}$, $k \neq l$    
    \State Add \emph{hard} pair $(x_i^{A_{k_*}}, x_i^{A_{l_*}})$ that maximizes 
    \Statex \hspace{1em} Eq. \ref{eq:acl} to the new batch with only hard pairs
\EndFor
    %\State Create minibatch $(\textbf{x}^{A_{k_*}}, \textbf{x}^{A_{l_*}})$ consisting of hard pairs 
    \State Proceed with standard SSL training
    %\State Update \(\theta\) with the hard batch 
    %Proceed with standard discriminiative SSL training to update 
    \State Repeat for all batches
\State \textbf{return} Pretrained model \(f\)
\label{alg}
\end{algorithmic}
\end{algorithm}
\end{minipage}
\end{wrapfigure}
Overall, we first generate $N$ augmented views for each image $x_i$ in the minibatch. Then, we forward these augmented views through the networks and create all combinatorially possible $\binom{N}{2}$ pairs of augmented images. Subsequently, we use Eq. \ref{eq:acl} to compute the sample-wise loss for each pair. We then select all pairs that yielded the highest loss to form the new \emph{hard} minibatch of augmented sets $\textbf{x}^{A_{k_*}}$ and $\textbf{x}^{B_{l_*}}$, discard the other pairs and use the hard minibatch for optimization. As shown in Algorithm \ref{alg}, we repeat this process in each training iteration. 

Intuitively, our approach introduces a more challenging learning scenario in which the model is encouraged to learn more discriminate features by being exposed to harder views. In the early stage of training, the embedding space lacks a defined structure for representing similarity among views. As training progresses, our method refines the concept of similarity through exposure to views that, from the perspective of the model, remain challenging given its current state. By limiting the number of sampled views, we upper-bound the difficulty of learning to prevent tasks from becoming too difficult and hindering learning. This ensures a controlled evolution of the embedding space, where the model's perception of difficulty is continuously challenged in tandem with its growing capacity to differentiate views. Consequently, \MethodAbbr{} can be seen as a regularization that prevents the model from overfitting to easy views.

While we exemplified the integration of \MethodAbbr{} with the SimSiam objective, integrating it into other contrastive methods is as straightforward. The only requirement of \MethodAbbr{} is to be able to compute sample-wise losses (to select the views with the highest loss). In our experiments section and in addition to SimSiam, we study the application of \MethodAbbr{} to the objectives of DINO, iBOT, and SimCLR (see also Appendix \ref{appendix:simclr_objective} for a formal exemplary definition for the integration of \MethodAbbr{} into SimCLR).

\section{Implementation and Evaluation Protocols}

\subsection{Implementation}

% what to mention here
% we do not checkpoint the forwards of ACL but could be done to improve time complexity
% write somewhere that, in order to reduce time complexity, we also looked at taking earlier activations 
We now describe the technical details of our approach. 
% TODO: Think about whether below sentence can be removed if too long?
\MethodAbbr{} can be used with any SSL method that allows computing sample-wise losses, and the only two elements in the pipeline we adapt are: 1) the data loader (which now needs to sample $N$ views for each image) and 2) the forward pass (which now invokes a \emph{select} function to identify and return the hard views). 
The image transformation distribution $\mathcal{T}$ taken from the baselines is left unchanged. Note, for the view selection one could simply resort to random resized crop (RRC) only and apply the rest of the operations after the hard view selection (see Section \ref{sec:what_makes_selection_hard} for a study on the influence of appearance on the selection).

%\textcolor{red}{To limit the cost of the additional forward passes required by \MethodAbbr{}, we optimized computational efficiency by reducing the image resolution by 50\% for these passes. Note, that for all other steps, such as the backward pass, we use the full-resolution images. Experiments have shown that this approach does not result in the selection of pairs less conducive to hard pretraining (see Appendix \ref{appendix:time_complexity} for more details).}

%and includes geometric (cropping \& resizing (RRC) and flipping) and appearance (color distortion, etc.) operations. 

All experiments were conducted with $N=4$ sampled views, yielding $\binom{N}{2}=6$ pairs to compare, except for DINO which uses 10 views (2 global, 8 local heads) per default. For DINO, we apply \MethodAbbr{} to both global and local heads but to remain tractable, we upper-bound the number of total pair comparisons to 128. SimCLR uses positive and negative samples. %and summarizes two loss terms, one for the positive pair and one for the negative pairs in the minibatch. 
Following the simplicity of \MethodAbbr{}, we do not alter its objective, which naturally leads to selecting hard views that are adversarial to positive and ``cooperative" to negative views. For iBOT, we use the original objective as defined by the authors with global views only.
\begin{wraptable}[22]{r}{0.54\linewidth}
\centering
\small
\tabcolsep=0.10cm
% \vspace{0.34cm}
% \vspace{-0.1cm}
\begin{tabular}{@{}llrrrr@{}}
\toprule
\multirow{2}{*}{\textbf{Method}} &
  \multicolumn{1}{c}{\multirow{2}{*}{\textbf{Arch.}}} &
  \multicolumn{2}{c}{\textbf{100 epochs}} &
  \multicolumn{2}{c}{\textbf{300 epochs}} \\ \cmidrule(l){3-6} 
 &
  \multicolumn{1}{c}{} &
  \multicolumn{1}{c}{Lin.} &
  \emph{k}-NN &
  \multicolumn{1}{c}{Lin.} &
  \emph{k}-NN \\ \midrule
DINO & ViT-S/16 & 73.52 & 68.80 & 75.48 & 72.62 \\
+ \MethodAbbr{} & ViT-S/16 & 74.67 & 70.72 & 76.56 & 73.65 \\
\textbf{Impr.} & & \textbf{+1.15} & \textbf{+1.92} & \textbf{+1.08} & \textbf{+1.03} \\ \midrule
DINO & RN50 & 71.93 & 66.28 & 75.25 & 69.53 \\
+ \MethodAbbr{} & RN50 & 72.87 & 67.33 & 75.65 & 70.05 \\
& & \textbf{+0.94} & \textbf{+1.05} & \textbf{+0.40} & \textbf{+0.52} \\ \midrule
SimSiam & RN50 & 68.20 & 57.47 & 70.35 & 61.40 \\
+ \MethodAbbr{} & RN50 & 68.98 & 58.97 & 70.90 & 62.97 \\
& & \textbf{+0.78} & \textbf{+1.50} & \textbf{+0.55} & \textbf{+1.57} \\ \midrule
SimCLR & RN50 & 63.37 & 52.83 & 65.50 & 55.65 \\
+ \MethodAbbr{} & RN50 & 65.33 & 54.76 & 67.30 & 56.80 \\
& & \textbf{+1.96} & \textbf{+1.93} & \textbf{+1.80} & \textbf{+1.15} \\ \midrule
iBOT & ViT-S/16 & 69.55 & 62.93 & 72.76 & 66.92 \\
+ \MethodAbbr{} & ViT-S/16 & 70.27 & 62.75 & 73.99 & 67.16 \\
& & \textbf{+0.73} & \textbf{-0.18} & \textbf{+1.23} & \textbf{+0.24} \\
\bottomrule
\end{tabular}
\caption{Average top-1 linear and \emph{k}-NN classification accuracy on the ImageNet validation set for 100 and 300-epoch pretrainings across 3 seeds.}
\label{tab:main_results}
\end{wraptable}

\vspace{-0.5cm}
\subsection{Evaluation Protocols}
We now describe the protocols used to evaluate the performance in our main results section. In self-supervised learning, it is common to assess pretraining performance with the linear evaluation protocol by training a linear classifier on top of frozen features or finetuning the features on downstream tasks. 
Our general procedure is to follow the baseline methods as closely as possible, including hyperparameters and code bases (if reported). It is common to use RRC and horizontal flips during training and report the test accuracy on central crops. Due to the sensitivity of hyperparameters, and as done by \cite{caron-iccv21a}, we also report the quality of features with a simple weighted nearest neighbor classifier (k-NN).

\section{Main Results}
Here, we discuss our main results on image classification, object detection, and segmentation tasks. All results are self-reproduced using the original baseline code and hyperparameters (see Appendix Section \ref{appendix:reproducibility_statement} for details). 

\subsection{Evaluations on ImageNet}
\label{sec:evaluations}

We report the top-1 validation accuracy on frozen features, as well as the k-NN classifier performance, in Table \ref{tab:main_results}. For DINO, we additionally compare ResNet-50~\citep{he-cvpr16a} against the ViT-S/16 \citep{dosovitskiy-iclr21a} architecture. We point out that both methods, vanilla, and vanilla+\MethodAbbr{} always receive the same number of data samples for training. Our method compares favorably against all baselines with an increased performance of approximately $1\%$ on average for 100 and 300 epoch pretraining, showing the benefit of sampling hard views.

Due to limited computing resources, we run the majority of pretrainings in this paper for 100 epochs and 300 epochs (200 epochs for SimCLR) and batch sizes of 512 (100 epoch) or 1024 (200 \& 300 epoch trainings), respectively. This choice is in line with a strategy that favors the evaluation of a diverse and larger set of baselines over the evaluation of a less diverse and smaller set and underpins the broad applicability of \MethodAbbr{}. 
We primarily ran our experiments with 8xNVIDIA GeForce RTX 2080 Ti nodes, with which the pretraining and linear evaluation duration ranged from ${\sim}3.5$ to ${\sim}25$ days. While \MethodAbbr{} requires roughly 2x the training time of the DINO baseline (see Appendix \ref{appendix:time_complexity} for further discussion on the time complexity of \MethodAbbr{}), this investment translates directly into improved performance. 

Rather than prioritizing efficiency, our focus was on achieving state-of-the-art results, highlighting \MethodAbbr{}'s flexibility and robustness across training durations. To showcase this point, we explored the scalability of \MethodAbbr{} with larger models and extended training schedules and achieved a new \textbf{state-of-the-art result of 78.8\% accuracy in linear evaluation, improving over the officially reported baseline of 78.2\% on DINO ViT-B/16 (400 epochs)}. For k-NN classification, the same model similarly surpassed the DINO baseline by 0.85\%, reaching 76.95\% compared to 76.1\%. These results demonstrate the scalability of our method to larger models and longer pretraining schedules. We emphasize that \MethodAbbr{} is insensitive to the baseline hyperparameters and simply reusing the default ones consistently resulted in improvements in the reported magnitudes across all experiments.

\subsection{Transfer to Other Datasets and Tasks}

We now report the transferability of features learned with \MethodAbbr{}. For all experiments here, we use our 100-epoch ImageNet pretrained iBOT and DINO ViT-S/16 models, respectively.

% \usepackage{booktabs}
% \usepackage{multirow}
% \usepackage{siunitx} % Add this package for better number alignment
\begin{table}[t]
\vspace{-0.5cm}
\centering
% \small
% \footnotesize
\tabcolsep=0.1cm
% \scriptsize
% \fontsize{8.5pt}{9pt}%\selectfont % Adjusts the font size to 7.5pt with a baselineskip of 9pt
\resizebox{\textwidth}{!}{%
\begin{tabular}{@{}lccccccccccc@{}}
\toprule
\multirow{2}{*}{\textbf{Method}} &
  \multirow{2}{*}{\textbf{Arch.}} &
  \multicolumn{2}{c}{\textbf{CIFAR10}} &
  \multicolumn{2}{c}{\textbf{CIFAR100}} &
  \multicolumn{2}{c}{\textbf{Flowers102}} &
  \multicolumn{2}{c}{\textbf{iNat 21}} &
  \multicolumn{2}{c}{\textbf{Food101}} \\ \cmidrule(lr){3-4} \cmidrule(lr){5-6} \cmidrule(lr){7-8} \cmidrule(lr){9-10} \cmidrule(l){11-12}
 &
   &
  Lin. &
  F.T. &
  Lin. &
  F.T. &
  Lin. &
  F.T. &
  Lin. &
  F.T. &
  Lin. &
  F.T. \\ \midrule
SimSiam &
  RN50 &
  82.60 &
  95.50 &
  54.20 &
  77.20 &
  34.27 &
  56.40 &
  32.50 &
  60.30 &
  65.70 &
  83.90 \\
\multicolumn{1}{c}{+ \MethodAbbr{}} &
  RN50 &
  84.40 &
  96.10 &
  57.10 &
  78.20 &
  38.37 &
  58.90 &
  33.90 &
  60.90 &
  67.10 &
  84.70 \\ \midrule
\textbf{Impr.} &
  \multicolumn{1}{l}{} &
  \textbf{+1.80} &
  \textbf{+0.60} &
  \textbf{+2.90} &
  \textbf{+1.00} &
  \textbf{+4.10} &
  \textbf{+2.50} &
  \textbf{+1.40} &
  \textbf{+0.60} &
  \textbf{+1.40} &
  \textbf{+0.80} \\ \midrule
DINO &
  ViT-S/16 &
  94.53 &
  98.53 &
  80.63 &
  87.90 &
  91.10 &
  93.20 &
  46.93 &
  53.97 &
  73.30 &
  87.50 \\
\multicolumn{1}{c}{+ \MethodAbbr{}} &
  ViT-S/16 &
  95.13 &
  98.65 &
  81.27 &
  88.23 &
  92.07 &
  93.60 &
  49.03 &
  54.16 &
  74.13 &
  87.91 \\ \midrule
\textbf{Impr.} &
  \multicolumn{1}{l}{} &
  \textbf{+0.60} &
  \textbf{+0.12} &
  \textbf{+0.63} &
  \textbf{+0.33} &
  \textbf{+0.97} &
  \textbf{+0.40} &
  \textbf{+2.10} &
  \textbf{+0.19} &
  \textbf{+0.83} &
  \textbf{+0.41} \\ \bottomrule
\end{tabular}
}

\caption{\MethodAbbr{} compares favorably against models trained without it when fine-tuned (F.T.) to or linearly evaluated (Lin.) on other datasets (averaged over 3 seeds; 100-ep. preraining).}
\label{tab:transfer_classification_results}
\vspace{-0.2cm}
\end{table}

\subsubsection{Linear Evaluation and Finetuning}
\begin{wraptable}[11]{r}{0.5\linewidth}
\vspace{-0.35cm}
% \fontsize{8pt}{9pt}%\selectfont
% \footnotesize
% \centering
% \tabcolsep=0.1cm
\resizebox{\linewidth}{!}{%
\begin{tabular}{@{}lccccc@{}}
\toprule
\textbf{Method} & \textbf{Arch.} & \multicolumn{2}{c}{\textbf{OD}} & \multicolumn{2}{c}{\textbf{IS}} \\
                &                & \textbf{100} & \textbf{300} & \textbf{100} & \textbf{300} \\
\midrule
iBOT            & ViT-S/16       & 66.13            & 66.80            & 63.10            & 63.63            \\
+ \MethodAbbr           & ViT-S/16       & 66.50            & 67.13            & 63.50            & 64.23            \\
\textbf{Impr.}         &                & \textbf{+0.37}            & \textbf{+0.33}            & \textbf{+0.40}            & \textbf{+0.60}            \\
\midrule
DINO            & ViT-S/16       & 65.90            & 66.60            & 62.83            & 63.63            \\
+ \MethodAbbr           & ViT-S/16       & 66.37            & 67.00            & 63.37            & 64.00            \\
\textbf{Impr.}         &                & \textbf{+0.47}            & \textbf{+0.40}            & \textbf{+0.53}            & \textbf{+0.37}            \\
\bottomrule

\end{tabular}
}
\caption{Object Detection (OD) and Instance Segmentation (IS) AP50 performance on COCO for 100/300 epoch pretraining.}
\label{tab:transfer_det_inst_seg}
\end{wraptable}

In Table \ref{tab:transfer_classification_results}, we apply both the linear evaluation (Lin.) and finetuning (F.T.) protocols to our models across a diverse set of datasets consisting of CIFAR10 \citep{krizhevsky-tech09a}, CIFAR100, Flowers102 \citep{nilsback-icvgip08a}, Food101 \citep{bossard-eccv14a}, and iNaturalist 2021 \citep{inaturalist21}. Our results show that the improvements achieved by sampling hard views that we observed so far also transfer to other datasets.

\subsubsection{Object Detection and Instance Segmentation}
For object detection and instance segmentation, we use the COCO \citep{lin-eccv14a} dataset with Cascade Mask R-CNN \citep{cai2019cascade, he-iccv17a}. Table \ref{tab:transfer_det_inst_seg}, where we report the AP50 performance, shows that the features learned with \MethodAbbr{} also transfer favorably to these tasks and outperform the iBOT and DINO baseline with a 100-ep. and 300-ep. pretraining. More details and performance results on this task are provided in Appendix \ref{appendix:add_results_od_is}.

\section{Empirical Analysis of \MethodAbbr{}}
\label{sec:empiricalanalysis}
In this section, we discuss studies designed to shed light on the mechanisms underlying \MethodAbbr{}. We address the following questions: 1) ``Which pattern can be observed that underlies the hard view selection?'' 
% 2) ``Can a ``manual" augmentation policy be inferred from these patterns?'', and 3
and 2) ``What are the effects on empowering the adversary?''. For all experiments conducted here, we use our 100-epoch, ImageNet-pretrained SimSiam+\MethodAbbr{} models with four sampled views. In Appendix \ref{appendix:manual_aug_policy}, we also analyze whether we can infer a ``manual" augmentation policy from the following observed patterns.

\subsection{Q1: Which Patterns Can be Observed with \MethodAbbr{}?}
\label{sec:what_makes_selection_hard}
When visually studying examples and the views selected by \MethodAbbr{} in Figures \ref{fig:example_views} and \ref{fig:examples_full_page} (in the appendix), we notice that both geometric and appearance characteristics seem to be exploited, for instance, see the brightness difference between the views of the first two rows in Fig. \ref{fig:example_views}.
%Note, that this observation is not limited to the examples depicted and noticed across all examples when applying \MethodAbbr{} to them. 
We also see a generally higher training loss (Fig. \ref{fig:training_loss_dino} in the appendix) indicative of an increased task difficulty. 

\begin{figure}
    \centering
    \begin{minipage}{0.49\linewidth} % Adjust the width as needed
        \centering
        \includegraphics[width=\linewidth]{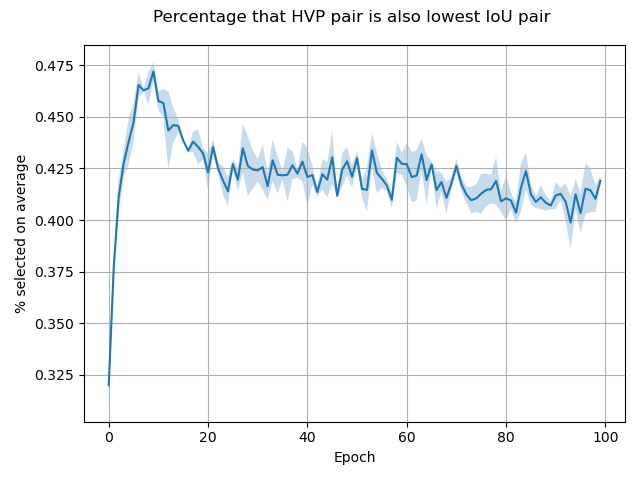}
    \end{minipage}
    %\hspace{0.1\linewidth} % Adjust the horizontal space between the two figures
    \begin{minipage}{0.49\linewidth} % Adjust the width as needed
        \centering
        \includegraphics[width=\linewidth]{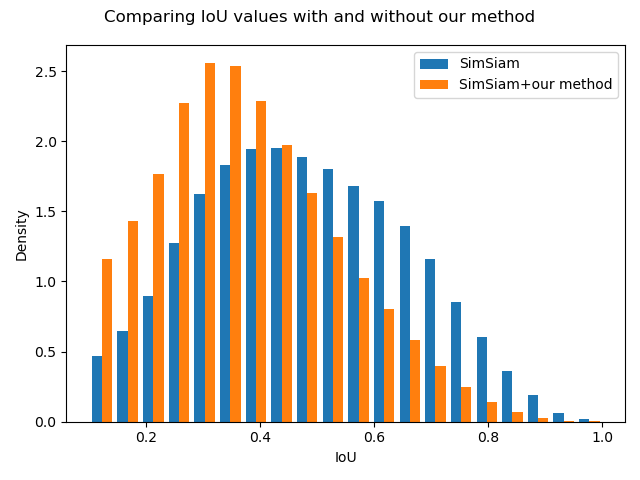}
    \end{minipage}
    \caption{\textbf{Left:} In over 40\% of the cases, the adversarially selected view pair has also the lowest Intersection over Union throughout SimSiam+\MethodAbbr{} pretraining. We attribute the early spike to the random initialization of the embedding. \textbf{Right:} \MethodAbbr{} (blue) shows a shift to smaller IoU values over standard pretraining (orange). Both results are based on 3 seeds.}
    \label{fig:iou_analysis}
\end{figure}

\subsubsection{Logging Augmentation Data}
To assess these observations, we logged relevant hyperparameter data during SimSiam training with \MethodAbbr{}. The logs include for each view the sampled geometric and appearance parameters from the data augmentation operations (such as the height/width of the crops or the brightness; see Section \ref{appendix:fanova} in the appendix for more details), as well as the loss and whether the view was selected. As evaluated metrics, we chose the Intersection over Union (IoU), Relative Distance (normalized distance of the center points views), color distortion distance (the Euclidean distance between all four color distortion parameters), and the individual color distortion parameters.
%i.e. all random resized crop parameters (height/ width of the original image, coordinates and height/width of crops); all Colorjitter strengths (brightness, contrast, saturation, hue); whether grayscale, Gaussian blurring, or horizontal flip was used; as well as the loss and if the crop was selected or not. As evaluated metrics we chose the Intersection over Union (IoU), Relative Distance (image-wise normalized distance of the center points of crop pairs), color distortion distance (the Euclidean distance between all four color distortion operation parameters), and the individual Colorjitter strengths.

\subsubsection{Importance of Augmentation Metrics} 
Given 300k such samples, we then used fANOVA \citep{hutter-icml14a} to determine how predictive these metrics are. This resulted in the metric with the highest predictive capacity on the loss being the IoU, explaining 15\% of the variance in performance, followed by brightness with 5\% (for more details see Fig. \ref{fig:fanova} in the appendix). The importance of IoU in \MethodAbbr{} is further underpinned by the following observation: the fraction of view pairs selected by \MethodAbbr{}, which are also the pairs with the lowest IoU among all six pairs (N=4), is over 40\% (random: ${\sim}16.7\%$) during training. Moreover, when using \MethodAbbr{}, a shift to smaller IoU values can be observed when comparing against standard pretraining (see Fig. \ref{fig:iou_analysis}). 

\begin{figure*}[t]
    \centering
    \begin{minipage}{0.49\linewidth} % Adjust the width as needed
        \centering
        \includegraphics[width=\linewidth]{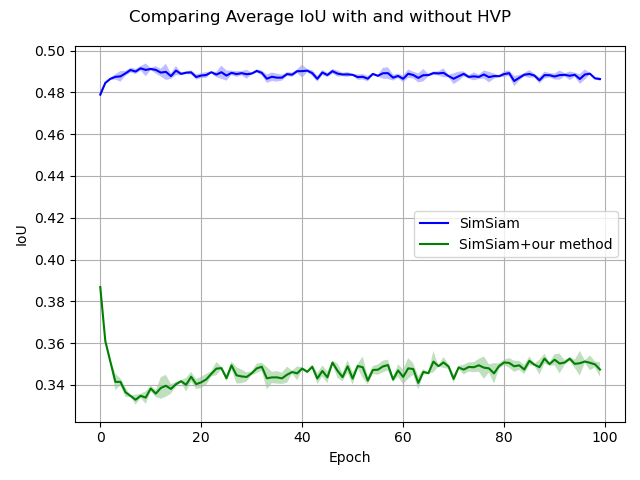}
    \end{minipage}
    %\hspace{0.1\linewidth} % Adjust the horizontal space between the two figures
    \begin{minipage}{0.49\linewidth} % Adjust the width as needed
        \centering
        \includegraphics[width=\linewidth]{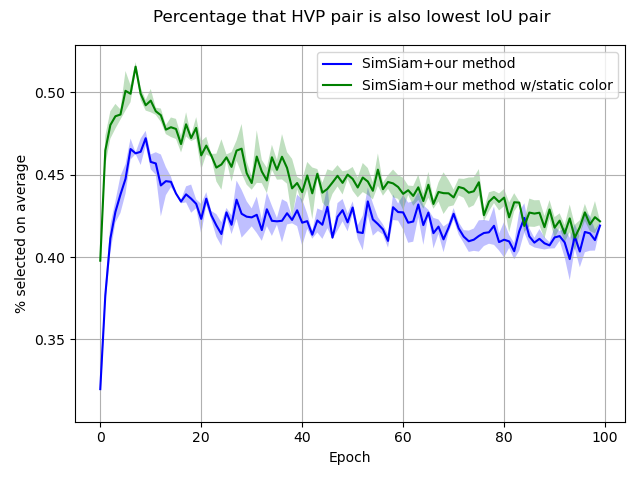}
    \end{minipage}
    \caption{\textbf{Left:} The average IoU of view pairs selected by SimSiam+\MethodAbbr{} (blue) compared against the default SimSiam training (green). \textbf{Right:} Using static color augmentation for all pairs before the selection increases the dependency on the IoU.}
    \label{fig:iou_analysis2}
\end{figure*}

\subsubsection{Taking a Closer Look at the Intersection over Union} 
We also examined the IoU value over the course of training in Fig. \ref{fig:iou_analysis2} (left). 
%Here, we notice the average IoU value of selected view pairs increases, possibly as a consequence of the pretrained model's embedding representation to the hard view selection.
%indicating that the pretrained model's embedding space becomes progressively better suited for identifying similarity of hard views. 
An observable pattern is that the IoU value with \MethodAbbr{} (Fig. \ref{fig:iou_analysis2} (left) in green) is smaller and varies more when compared against training without \MethodAbbr{} (blue). We believe this is due to the sample-wise and stateful nature of the adversarial selection as \MethodAbbr{} chooses different IoU values between varying samples and model states. 

Lastly, we assessed the effect of the color augmentation on the pair selection. For this study, we sampled \emph{one} set of color augmentations (as opposed to one for each view) per iteration and applied it to all views.  We apply sampled data augmentations to each view only after identifying the hardest pair.  
%Consequently, the linear evaluation performance dropped by 0.3\% on average. 
As we show in Fig. \ref{fig:iou_analysis2} (right), the fraction of selected pairs that are also the hardest pairs slightly increases in this case. One possible explanation for this is that it reflects the non-negligible role of color variation between views (as shown previously with the importance analysis), where \MethodAbbr{} is given less leverage to increase hardness through a static appearance and instead, depends more on leveraging the IoU. Another key observation is that \MethodAbbr{} often chooses view pairs that incorporate zooming in and out or an increased distance between the views (see last row of Fig. \ref{fig:example_views}). %We believe the former relates to the analogy of local-to-global view correspondence described in SimCLR and the DINO papers.

\subsection{Q2: What are the Effects of Strong Adversaries?}
It is well known that adversarial learning can suffer from algorithmic instability \citep{xing-neurips21a}, e.g. by giving an adversary too much capacity. Here, we further explore the space of adversarial capacity for pretraining with hard views by adapting and varying \MethodAbbr{} hyperparameters in order to gain a better understanding of their impact and robustness on discriminative learning. Additionally, we report further results on learning an adversary in Appendix \ref{appendix:stn}. 

\subsubsection{Robustness to Augmentation Hyperparameters} 
\begin{figure}[t]
    \vspace{-1cm}
    \centering
    \begin{minipage}{0.49\linewidth}
        \centering
        \includegraphics[width=\linewidth]{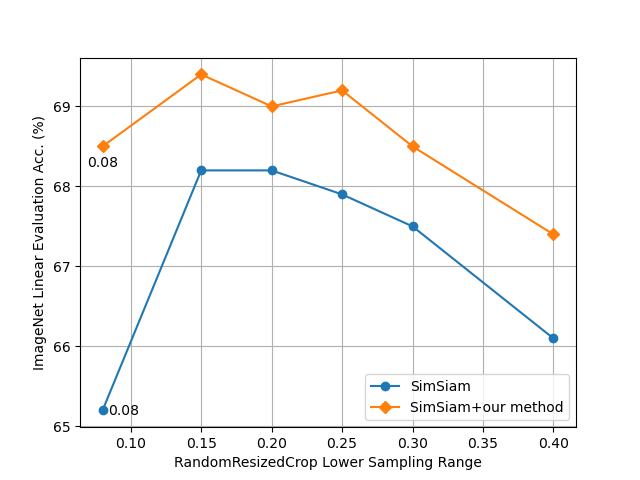}
    \end{minipage}
    %\hspace{0.1\linewidth}
    \begin{minipage}{0.49\linewidth}
        \centering
        \includegraphics[width=\linewidth]{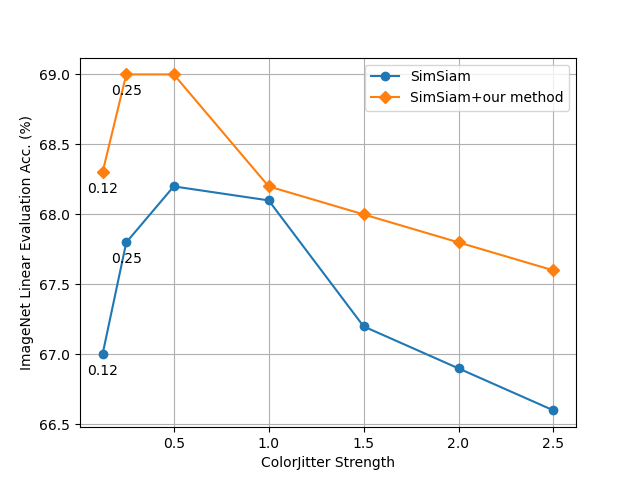}
    \end{minipage}
    \caption{With \MethodAbbr{}, SimSiam appears more robust to augmentation hyperparameter variation. We show this for RandomResizedCrop (\textbf{left}) and ColorJitter (\textbf{right}). For RRC, the values indicate the lower value of the sampling range and for CJ the intensity of the color cues. Results averaged over two seeds and SimSiam defaults are RRC=0.2 and CJ=0.5.}
    \label{fig:robustness}
\end{figure}

In assessing the robustness of \MethodAbbr{} to augmentation hyperparameters, we conducted an analysis focusing on two primary augmentation operations: RandomResizedCrop (RRC) and ColorJitter (CJ). Our findings depicted in Fig. \ref{fig:robustness} suggest that \MethodAbbr{} enhances the robustness of SSL methods, like SimSiam, against variations in these hyperparameters. Note, when varying one operation, either RRC or CJ, we maintained the other operation at its default configuration. In both settings, we observe less performance degradation with extreme augmentation values and overall smaller degradation rates for \MethodAbbr{}. We believe that this robustness stems from hard view pretraining which inadvertently equips the model to handle stronger augmentations.

\subsubsection{Increasing the Number of Views} In our initial experiments, we explored variations in the number of sampled views $N$ with SimSiam and \MethodAbbr{}. As can be seen in Fig. \ref{fig:more_views} in the appendix, while $N=8$ views still outperform the baseline in terms of linear evaluation accuracy, it is slightly worse than using $N=4$ views (-0.05\% for 100 epochs
%and -0.14\% for 200 epochs
pretraining on linear evaluation). We interpret this result as the existence of a ``sweet spot" in setting the number of views, where, in the limit, a higher number of views corresponds to approximating a powerful adversarial learner, capable of choosing very hard and unfavorable learning tasks that lead to model collapse and performance deterioration. We experimented with such an adversarial learner and report results in Appendix \ref{appendix:stn}.
%\footnote{We also experimented with a cooperative view generator (see Appendix \ref{appendix:easy_view_selection}).}
\section{Conclusion}
We presented \MethodAbbr{}, a new data augmentation and learning strategy for Self-Supervised Learning designed to challenge pretrained models with harder samples. This straightforward method allows pushing the effectiveness of the traditional random view generation in SSL. When combined with methods like DINO, SimSiam, iBOT, and SimCLR, \MethodAbbr{} consistently showcased improvements of 1\% on average in linear evaluation and a diverse set of transfer tasks. \MethodAbbr{} achieved a new state-of-the-art result of 78.8\% linear evaluation accuracy on DINO ViT-B/16, a 0.6\% improvement over the previous baseline. This illustrates the scalability and effectiveness of our approach for larger pretraining settings. With growing models, there is an increasing demand for more data to effectively train them. Synthetic data generation offers a viable solution by enhancing the quantity and diversity of training data. Data augmentation techniques like \MethodAbbr{}  play a crucial role in this process by creating challenging views, which can serve as synthetic data. All in all, \MethodAbbr{} holds promise for scenarios where one seeks to push absolute performance or explore making models less sensitive to hyperparameters, thereby strengthening them for various downstream applications.
\section{Acknowledgements}
Frank Hutter acknowledges the financial support of the Hector Foundation. We also acknowledge funding by the European Union (via ERC Consolidator Grant DeepLearning 2.0, grant no.~101045765). Views and opinions expressed are however those of the author(s) only and do not necessarily reflect those of the European Union or the European Research Council. Neither the European Union nor the granting authority can be held responsible for them. Moreover, we acknowledge funding by the Deutsche Forschungsgemeinschaft (DFG, German Research Foundation) under grant number 417962828, by the state of Baden-W\"urttemberg through bwHPC and the German Research Foundation (DFG) through grant INST 35/1597-1 FUGG, as well as the Gauss Center for Supercomputing eV (www.gauss-centre.eu) for funding this project by providing computing time on the GCS supercomputer JUWELS at J\"ulich Supercomputing Center (JSC). We also acknowledge Robert Bosch GmbH for financial support and the computing time made available on the high-performance computer NHR@KIT Compute Cluster at the NHR Center NHR@KIT.

\begin{center}\includegraphics[width=0.3\textwidth]{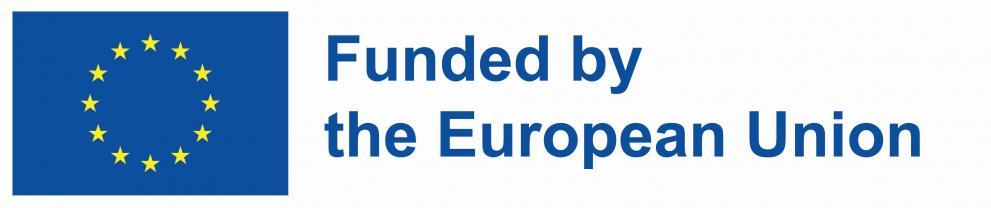}\end{center} 

% \bibliography{iclr2025_conference}
\bibliographystyle{iclr2025_conference}
\bibliography{temp, lib, shortproc, shortstrings}

\newpage
\appendix

\appendix
\section*{Appendix}

\section{Reproducibility Statement}
\label{appendix:reproducibility_statement}
We provide complete code, environment installation instructions, hyperparameter settings and model checkpoints here: \url{https://github.com/automl/hvp}. For transparency, we outline all hyperparameters, data splits, and evaluation protocols in detail in Section \ref{appendix:hypers}. Most experiments were run across multiple seeds, and we report average results to account for variability. Information regarding the required computational resources is discussed in Section \ref{appendix:time_complexity} below.

\section{Examples Sampled by \MethodAbbr{}}

\begin{figure}[h!]
    \centering
    \includegraphics[width=0.5\linewidth]{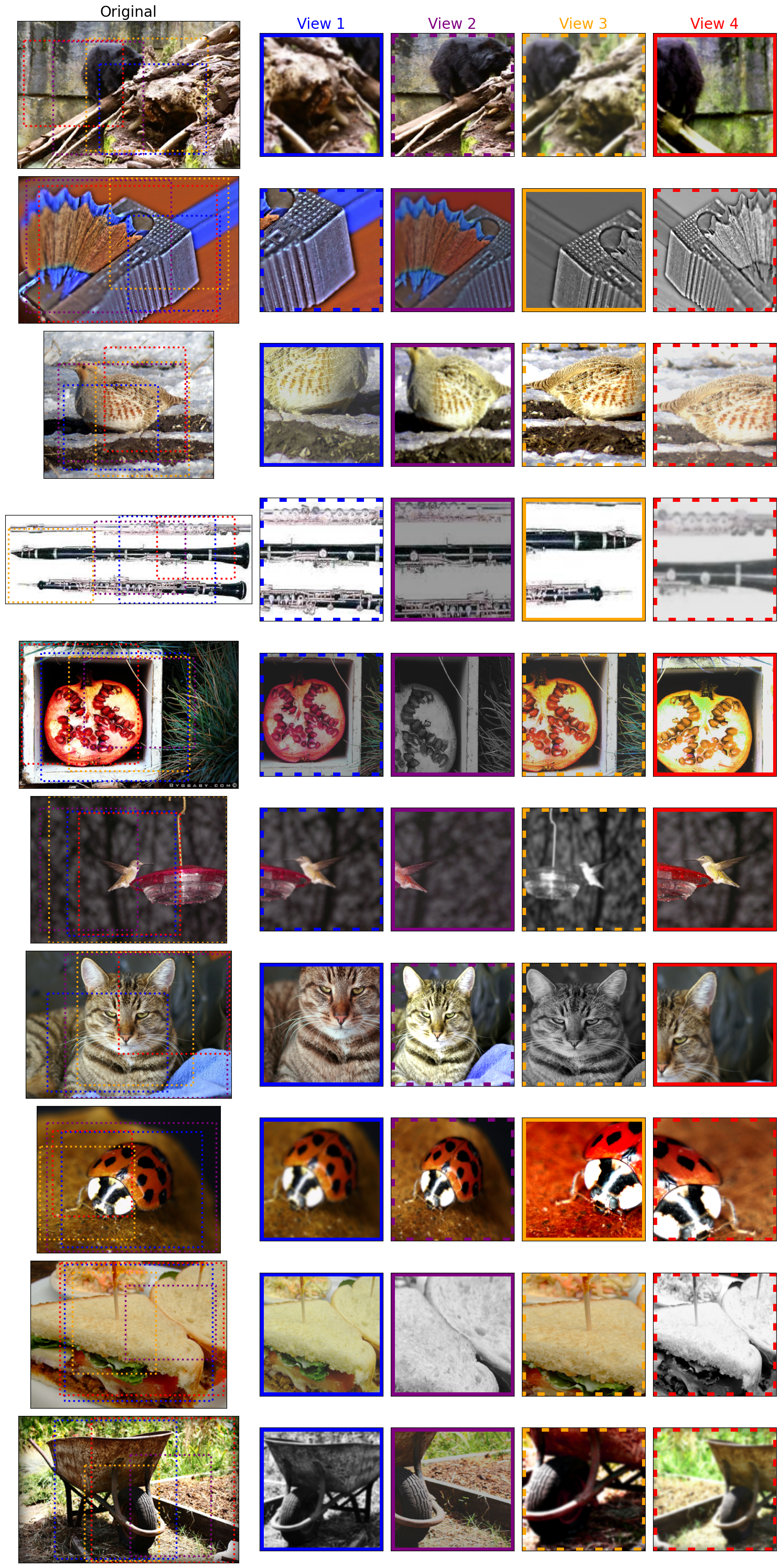}
    \caption{We depict row-wise ten example images from the ImageNet train set along with their sampled views with a finished, 100-epoch trained SimSiam ResNet50 model. Left: original image with the overlaid randomly sampled crops (colored dashed rectangles). Right: All views after applying resizing and appearance augmentations. The pair that is selected adversarially by \MethodAbbr{} is highlighted in solid lines, eg. View 1 and View 4 in the first row.}
    \label{fig:examples_full_page}
\end{figure}

\section{Training Loss}
\begin{figure}[H]
    \centering
        \includegraphics[width=0.5\linewidth]{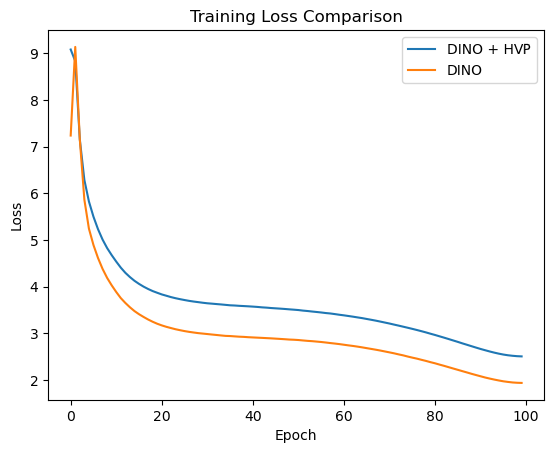}
    \caption{The training loss over 100 epochs. Comparing the DINO vanilla method with DINO + \MethodAbbr{}. The spike and drop in the loss curve of DINO is caused by freezing the last layer in the first epoch which was proposed by the authors as a strategy to enhance downstream performance. For \MethodAbbr{} we can only see a drop and no spike. We believe this is because \MethodAbbr{} exposes the model to hard views from the beginning of training (i.e. the loss is immediately maximized).}
    \label{fig:training_loss_dino}
\end{figure}

\section{Effect of More Views on Linear Evaluation Performance}
\begin{figure}[H]
    \centering
        \includegraphics[width=\linewidth]{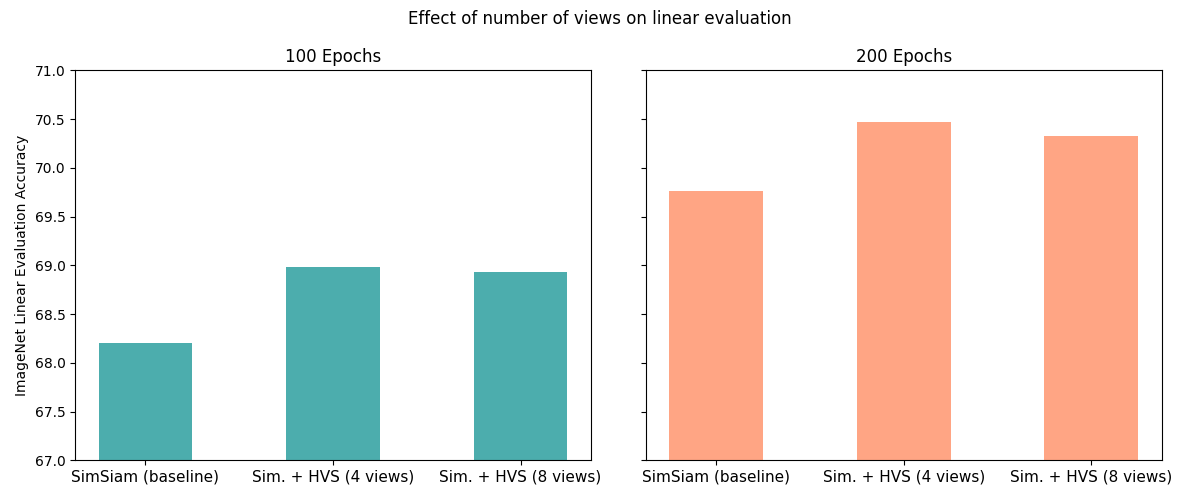}
    \caption{Setting the number of views too high can result in performance deterioration. This shows that diminishing returns exist, likely because the adversary becomes too strong, resulting in a too hard learning task.}
    \label{fig:more_views}
\end{figure}

\section{Assessing the Importance of Metrics with fANOVA}
\label{appendix:fanova}
\begin{figure}[H]
    \centering
        \includegraphics[width=0.8\linewidth]{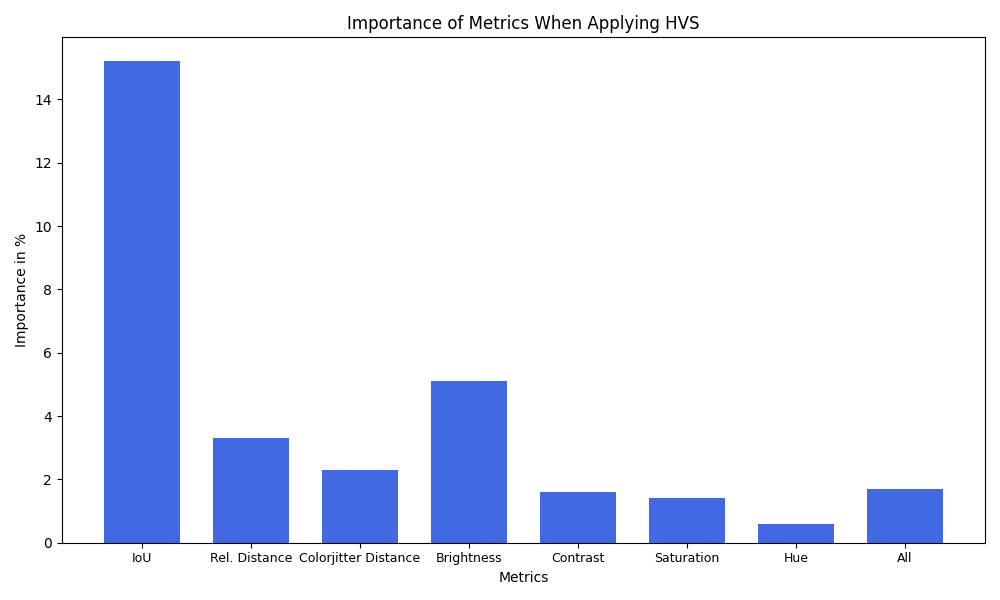}
    \caption{Application of fANOVA \cite{hutter-icml14a} on logged training data to determine metrics with high predictive capacity on the training loss.}
    \label{fig:fanova}
\end{figure}

To assess the importance of various metrics on the training loss, we apply fANOVA \cite{hutter-icml14a} on data that we logged during training with \MethodAbbr{}. We used 300k samples that contain the following sampled parameters from the geometric and appearance data augmentation operations for each view: all random resized crop parameters (height and width of the original image, coordinates of crop corners and height and width of the crop), all Colorjitter (color distortion) strengths (brightness, contrast, saturation, hue), grayscale on/off, Gaussian blurring on/off, horizontal flip on/off, loss, and if the crop was selected or not. The metrics we chose are Intersection over Union (IoU), Relative Distance (sample-wise normalized distance of the center points of crop pairs), color distortion distance (the Euclidean distance between all four color distortion operation parameters, i.e. brightness, contrast, saturation, hue), and the individual color distortion parameters of the Colorjitter operation. As can be seen in Fig. \ref{fig:fanova}, the metric with the highest predictive capacity on the loss is the IoU with an importance of 15.2\% followed by brightness with 5.1\%. The relative distance has an importance of 3.3\%, the Colorjitter distance 2.3\%, the contrast 1.6\%, the saturation 1.4\%, hue 0.6\%, and all parameters jointly 1.7\%.

\section{Additional Empirical Analysis}
\subsection{Can a Manual Augm. Policy be Derived?}
\label{appendix:manual_aug_policy}
% \begin{wraptable}[9]{}{0.45\linewidth}
\begin{table}%[12]{R}{0.43\linewidth}
\centering
\small
\vspace{-.8\intextsep}
\begin{tabular}{lll}
\toprule
\textbf{IoU Policy Type}               & \textbf{SimSiam} & \textbf{DINO} \\ \midrule
Baseline (B)                          & 68.20          & 73.50       \\
B+range(0.3-0.35)               & -0.80          & -1.47      \\
B+range(0.3-0.35)+alt. & +0.10          & -0.45       \\
B+range(0.4-0.6)                & +0.55          & -0.40      \\
B+range(0.4-0.6)+alt.  & +0.25          & -0.20       \\
B+range(0.1-0.8)                & -33.95         & -1.50       \\
B+range(1.0-0.1)        & +0.07            & -             \\ \bottomrule
\end{tabular}
\caption{Top-1 lin. eval. accuracies for the manual IoU policy.}
\label{tab:iou_policy}
\end{table}
% \end{wraptable}

Since harder pretraining tasks seem beneficial according to observations made in Q1, a natural question arises: can we mimic the adversarial selection with a manually scripted augmentation policy? Such a policy would replace \MethodAbbr{} and lower the computational cost. Since the IoU plays an important role, below, we study several ways to construct a simple manual augmentation policy based on IoU.

%\vspace{-0.2cm}
\subsubsection{Deriving an Augmentation Policy} We implemented the following rejection sampling algorithm in the augmentation pipeline: we linearly approximate the IoU values from Fig. \ref{fig:iou_analysis2} (left; in blue) with start and end values of 0.30 and 0.35 (ignoring the dip in the early phase). For each iteration, we then check if the pairs exceed the IoU value and if so, we reject the pair and re-sample a new pair. This ensures that only pairs are sampled that entail a minimum task difficulty (by means of a small enough IoU). We varied different hyperparameters, e.g., IoU start/end ranges, inverse schedules, and alternating between the IoU schedule and the standard augmentation
%every other iteration; as well as warmup, i.e. no IoU policy for the first 10 epochs)
. Training both SimSiam and DINO 
%(ViT-S/16; applied to all global and local heads) 
models for 100 epochs yielded performance drops or insignificant improvements (see Table \ref{tab:iou_policy}). 
%Moreover, we see an increased risk of model collapse when setting start values too small, which can be interpreted as a too hard task. 
These results indicate that using a manual policy based on metrics in pixel space such as the IoU is non-trivial. Additionally, these results show that transferring such a policy from SimSiam to DINO does not work well, possibly due to additional variations in the augmentation pipeline such as multi-crop. In contrast, we believe that \MethodAbbr{} is effective and transfers well since it 1) operates on a similarity level of latent embeddings that may be decorrelated from the pixel space and 2) has access to the current model state.

\subsubsection{Assessing the Difficulty of Predicting the Pair Selected by \MethodAbbr{}} 
We further validate the previous result and assess an upper limit on the performance for predicting the hardest pair. By fitting a gradient boosting classification tree \cite{friedman-as01a}, we predict the selected view pair conditioned on all SimSiam hyperparameter log data from Q1 except for the flag that indicates whether a view was selected. As training and test data, we used the logs from two seeds (each 300k samples) and the logs from a third seed, respectively. We also tuned hyperparameters on train/valid splits and applied a 5-fold CV. However, the resulting average test performance in all scenarios never exceeded 40\%, indicating that it is indeed challenging to predict the hardness of views based on parameter-level data. This further supports our hypothesis that deriving a policy for controlling and increasing hardness based on geometric and appearance parameters is non-trivial and that such a policy must function on a per-sample basis and have access to the current model state as in \MethodAbbr{}.

\section{Learning View Generation}
\subsection{Adversarial Learner}
\label{appendix:stn}
In this experiment, we explore adversarially learning a network to output the transformation matrix for view generation. We use Spatial Transformer Network (STN) \cite{jaderberg-nips15a} to allow generating views by producing 6D transformation matrices (allowing translating, rotating, shearing, scaling, affine transformations, and combinations thereof) in a differentiable way since most common augmentations are not off-the-shelf differentiable. We optimize the STN jointly with the DINO objective and a ViT-tiny. We train it alongside the actual pretrained network using the same (inverted) objective. For our experiments, we use DINO with multi-crop, i.e. 2 global and 8 local heads. As STN we use a small CNN followed by a linear layer for outputting the 10*6D transformation matrices. In this scenario, we use a ViT-tiny/16 with a 300 epoch pretraining on CIFAR10 with a batch size of 256. All other hyperparameters are identical to the ones reported in the DINO paper.

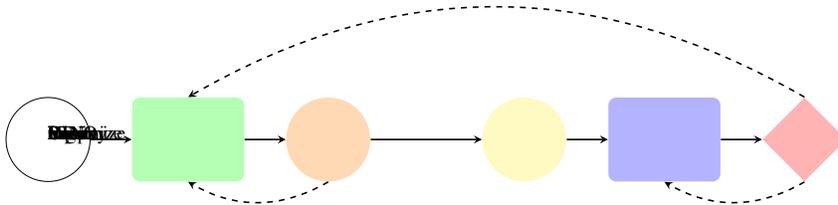
\begin{figure}[h]
    \centering
    \resizebox{0.8\linewidth}{!}{%
    \begin{tikzpicture}[node distance=2cm]
        \node (input) [crclwhite] {Input};
        \node (stn) [rectgreen, right of=input, xshift=0.5cm] {STN};
        \node (crops) [crclorange, right of=stn, xshift=0.5cm] {Crops};
        \node (views) [crclyellow, right of=crops, xshift=1.5cm] {Views};
        \node (dino) [rectblue, right of=views, xshift=0.5cm] {DINO};
        \node (loss) [dmnd, right of=dino, xshift=0.5cm] {Loss};
        
        \draw [arrow] (input) -- (stn);
        \draw [arrow] (stn) -- (crops);
        \draw [arrow] (crops) -- node[anchor=south] {augm.} (views);
        \draw [arrow] (views) -- (dino);
        \draw [arrow] (dino) -- (loss);
        \draw [arrow, dashed] (loss.north) to [bend right] node[anchor=south] {maximize} (stn.north);
        \draw [arrow, dashed] (loss.south) to [bend left] node[anchor=north] {minimize} (dino.south);
        \draw [arrow, dashed] (crops.south) to [bend left] node[anchor=north] {Penalty} (stn.south);
    \end{tikzpicture}%
    }
    \caption{Illustration of adversarial learning with a Spatial Transformer Network (STN) jointly with contrastive learning (here: DINO).}
    \label{fig:stn}
\end{figure}

Figure \ref{fig:stn} visualizes the procedure. The STN takes the raw image input and generates a number of transformation matrices that are applied to transform the image input into views. These views are then passed to the DINO training pipeline. Both networks are trained jointly with the same loss function. DINO is trained with its original contrastive objective, where the STN is trained by inverting the gradient after the DINO during backpropagation.

\begin{figure}[H]
    \centering
        \includegraphics[width=\linewidth]{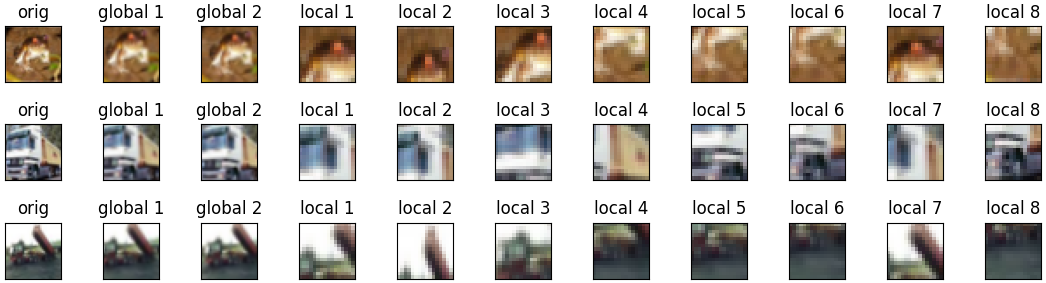}
    \caption{Example for static output behavior of the STN.}
    \label{fig:stn_examples}
\end{figure}

The STN, without using auxiliary losses, starts zooming in and generating single-color views. To counteract this behavior, we experimented with different penalties on the transformation matrices produced by the STN.  For instance, in order to limit the zooming pattern, we can use the determinants of the sub-matrices of the transformation matrix to penalize based on the area calculated and apply a regression loss (e.g. MSE). We refer to this type of penalty as \emph{Theta Crops Penalty} (TCP). Additionally, we also restrict its parameters to stay within a sphere with different parameters for local and global crops. Next to determinant-based penalty losses, we also experimented with other penalty functions such as the weighted MSE between the identity and the current transformation matrix or penalties based on histograms of the input image and generated views after applying the transformation. To avoid strong uni-dimensional scaling behavior, we also implemented restricting scaling in a symmetric way (i.e. applied to both x and y dimensions) and refer to this as \emph{scale-sym.}. We report our best results in Table \ref{tab:stn} which are all TCP-based. As can be seen, no setting is able to outperform the baseline. Our best score was achieved with translation-scale-symmetric which is very similar to random cropping. When removing the symmetries in scaling, the performance drops further. Removing a constraint adds one transformation parameter and therefore one dimension. This can be seen as giving more capacity to the adversarial learner which in turn can make the task significantly harder. Similarly, when adding rotation, the performance drops further and in part drastically. This is on the one hand due to the penalties not being fully able to restrict the output of the STN. On the other, the task of extracting useful information from two differently rotated crops is even harder, and learning spatial invariance becomes too challenging. All in all, we experienced two modes: either the STN is too restricted, leading to \emph{static output} (i.e. independent of image content, the STN would produce constant transformation matrices) or the STN has too much freedom, resulting in extremely difficult tasks. See Fig. \ref{fig:stn_examples} for an example of the former behavior.
% DINO no multi-crop

\begin{table}[ht]
    \centering
    \begin{tabular}{lccc}
        \toprule
        Mode & Penalty & Lin. & F.T. \\ [0.25ex]
        \hline
        baseline & - & 86.1 & 92.7 \\ 
        translation-scale-sym. & TCP & 83.7 & 90.3 \\
        translation-scale & TCP & 82.8 & 89.7 \\
        rotation-translation & TCP & 56.7 & - \\
        rotation-translation-scale & TCP & 31.7 & - \\
        rotation-translation-scale-sym. & TCP & 77.6 & - \\
        affine & TCP & 78.3 & 83.5 \\
        \bottomrule
    \end{tabular}
    \caption{\textbf{Linear evaluation and finetuning classification performance on CIFAR10}. Top-1 accuracy on the validation set of CIFAR10 for our best results reported with different STN transformation modes.}
    \label{tab:stn}
\end{table}

\subsection{Cooperative Learner}
\label{appendix:easy_view_selection}
To investigate the effect of a \emph{cooperative}, i.e. easy pair selection, we conducted a small experiment. Instead of selecting the pair yielding the worst loss, we inverted the objective and selected the pair with the best loss. As expected, this led to model collapses with a linear eval. performance of 0.1\%. This result is in line with previous findings that highlight the importance of strong augmentations in CL.

\section{Additional Results}
\subsection{Object Detection and Instance Segmentation}
Here we provide more detailed results on object detection and instance segmentation, shown in Tab. \ref{tab:transfer_det_inst_seg_full} and Tab. \ref{tab:transfer_det_inst_seg_full_ep300} respectively. We followed iBOT's default configurations which employed Cascade Mask R-CNN as the task layer.
\label{appendix:add_results_od_is}
% \usepackage{booktabs}
\begin{table*}[h]
\centering
\begin{tabular}{@{}llllllll@{}}
\toprule
\textbf{Method} & \textbf{Arch.} & \multicolumn{3}{c}{\textbf{Object Detection}} & \multicolumn{3}{c}{\textbf{Instance Segmentation}} \\
\midrule
 & & $\mathbf{AP^b}$ & $\mathbf{AP^b_{50}}$ & $\mathbf{AP^b_{75}}$ & $\mathbf{AP^m}$ & $\mathbf{AP^m_{50}}$ & $\mathbf{AP^m_{75}}$ \\
\midrule
iBOT                      & ViT-S/16 &         47.00  &         66.13  &         50.63  &         40.67  &         63.10  &         43.37  \\
\multicolumn{1}{r}{+ \MethodAbbr} & ViT-S/16 &         47.27  &         66.50  &         50.90  &         40.90  &         63.50  &         43.83  \\
\textbf{Improvement}      &          & \textbf{+0.27} & \textbf{+0.37} & \textbf{+0.27} & \textbf{+0.23} & \textbf{+0.40} & \textbf{+0.47} \\
\midrule
DINO                      & ViT-S/16 &         46.50  &         65.90  &         50.30  &         40.43  &         62.83  &         43.27  \\
\multicolumn{1}{r}{+ \MethodAbbr} & ViT-S/16 &         47.07  &         66.37  &         50.63  &         40.80  &         63.37  &         43.87  \\
\textbf{Improvement}      &          & \textbf{+0.57} & \textbf{+0.47} & \textbf{+0.33} & \textbf{+0.37} & \textbf{+0.53} & \textbf{+0.60} \\
\bottomrule
\end{tabular}
\caption{Additional object detection and instance segmentation results on the COCO dataset. The ViT-S/16 models were pretrained for 100 epochs.}
\label{tab:transfer_det_inst_seg_full}
\end{table*}

\begin{table*}[h]
\centering
\begin{tabular}{@{}llllllll@{}}
\toprule
\textbf{Method} & \textbf{Arch.} & \multicolumn{3}{c}{\textbf{Object Detection}} & \multicolumn{3}{c}{\textbf{Instance Segmentation}} \\
\midrule
 & & $\mathbf{AP^b}$ & $\mathbf{AP^b_{50}}$ & $\mathbf{AP^b_{75}}$ & $\mathbf{AP^m}$ & $\mathbf{AP^m_{50}}$ & $\mathbf{AP^m_{75}}$ \\
\midrule
iBOT                      & ViT-S/16 &         47.60  &         66.80  &         51.33  &         41.10  &         63.63  &         44.07  \\
\multicolumn{1}{r}{+ \MethodAbbr} & ViT-S/16 &         48.03  &         67.13  &         51.73  &         41.50  &         64.23  &         44.40  \\
\textbf{Improvement}      &          & \textbf{+0.43} & \textbf{+0.33} & \textbf{+0.40} & \textbf{+0.40} & \textbf{+0.60} & \textbf{+0.33} \\
\midrule
DINO                      & ViT-S/16 &         47.27  &         66.60  &         51.00  &         41.00  &         63.63  &         44.03  \\
\multicolumn{1}{r}{+ \MethodAbbr} & ViT-S/16 &         47.50  &         67.00  &         51.33  &         41.17  &         64.00  &         44.13  \\
\textbf{Improvement}      &          & \textbf{+0.23} & \textbf{+0.40} & \textbf{+0.33} & \textbf{+0.17} & \textbf{+0.37} & \textbf{+0.10} \\
\bottomrule
\end{tabular}
\caption{Additional object detection and instance segmentation results on the COCO dataset. The ViT-S/16 models were pretrained for 300 epochs.}
\label{tab:transfer_det_inst_seg_full_ep300}
\end{table*}

\section{Hyperparameters}
\label{appendix:hypers}
\subsection{Evaluations on ImageNet}
\subsubsection{DINO}
For DINO, we report the ViT pretraining hyperparameters in Table \ref{tab:hypers_dino} (ViT-S) and Table \ref{tab:hypers_dino_vitb} (ViT-B) which are the original ones as reported by the authors. Note, for \MethodAbbr{} we limit the total number of comparisons to 128 across all heads. Linear evaluation is executed for 100 epochs and we use a learning rate of $0.00075$, SGD optimizer (AdamW \cite{loshchilov-iclr19a} during pretraining), a batch size of 1024, a momentum of $0.9$, and no weight decay (as reported by the authors).

% \usepackage{booktabs}
\begin{table*}[h!]
\centering
\begin{tabular}{@{}ll|ll@{}}
\toprule
\textbf{Hyperparameter}        & \textbf{Value} & \textbf{Hyperparameter} & \textbf{Value} \\ \midrule
architecture                   & vit-small     & epochs:                 & 100            \\
img-size                     & 224            & warmup-epochs:         & 10             \\
patch-size                   & 16             & freeze-last-layer:    & 1              \\
out-dim                      & 65536          & lr:                     & 0.0005         \\
norm-last-layer             & true           & min-lr:                & 1.0e-06        \\
momentum-teacher             & 0.996          & optimizer:              & AdamW          \\
use-bn-in-head             & false          & weight-decay:          & 0.04           \\
teacher-temp                 & 0.04           & weight-decay-end:     & 0.4            \\
warmup-teacher-temp         & 0.04           & global-crops-scale:   & 0.4, 1.0       \\
warmup-teacher-temp-epochs & 0              & global-crops-size:    & 224            \\
fp16                          & true           & local-crops-number:   & 8              \\
batch-size                   & 512            & local-crops-scale     & 0.05, 0.4      \\
clip-grad                    & 3.0            & local-crops-size:     & 96             \\
drop-path-rate              & 0.1            &                         &                \\ \bottomrule
\end{tabular}
\caption{Pretraining ImageNet hyperparameters for the runs with DINO ViT-S/16. For 300 epochs, we use a batch size of 1024.}
\label{tab:hypers_dino}
\end{table*}

\begin{table*}[h!]
\centering
\begin{tabular}{@{}ll|ll@{}}
\toprule
\textbf{Hyperparameter}        & \textbf{Value}  & \textbf{Hyperparameter} & \textbf{Value} \\ \midrule
architecture                   & vit-base        & epochs:                 & 400            \\
img-size                       & 224             & warmup-epochs:          & 10             \\
patch-size                     & 16              & freeze-last-layer:      & 3              \\
out-dim                        & 65536           & lr:                     & 0.00075        \\
norm-last-layer                & true            & min-lr:                 & 2.0e-06        \\
momentum-teacher               & 0.996           & optimizer:              & AdamW          \\
use-bn-in-head                 & false           & weight-decay:           & 0.04           \\
teacher-temp                   & 0.07            & weight-decay-end:       & 0.4            \\
warmup-teacher-temp            & 0.04            & global-crops-scale:     & 0.25, 1.0      \\
warmup-teacher-temp-epochs     & 50              & global-crops-size:      & 224            \\
fp16                           & false           & local-crops-number:     & 10             \\
batch-size                     & 1024            & local-crops-scale:      & 0.05, 0.25     \\
clip-grad                      & 0.3             & local-crops-size:       & 96             \\
drop-path-rate                 & 0.1             &                         &                \\ \bottomrule
\end{tabular}
\caption{Pretraining ImageNet hyperparameters for the runs with DINO ViT-B/16.}
\label{tab:hypers_dino_vitb}
\end{table*}

\subsubsection{SimSiam}
In Table \ref{tab:hypers_simsiam}, we report the ResNet-50 pretraining hyperparameters. Linear evaluation is executed for 90 epochs (as reported by the SimSiam authors) and we use a learning rate of 0.1, LARS optimizer \cite{lars}, a batch size of 4096, and no weight decay.
% \usepackage{booktabs}
\begin{table}[h!]
\centering
\begin{tabular}{@{}ll}
\toprule
\textbf{Hyperparameter} & \textbf{Value} \\ \midrule
architecture                   & resnet50      \\
batch-size            & 512           \\
blur-prob             & 0.5           \\
crops-scale            & 0.2, 1.0      \\
crop-size             & 224           \\
feature-dimension     & 2048          \\
epochs                 & 100           \\
fix-pred-lr          & true          \\
lr                     & 0.05       \\   
momentum               & 0.9      \\     
predictor-dimension    & 512    \\       
weight-decay          & 0.0001 \\
optimizer              & SGD \\
\bottomrule
\end{tabular}
\caption{Pretraining ImageNet hyperparameters for the runs with SimSiam. For 300 epochs, we use a batch size of 1024.}
\label{tab:hypers_simsiam}
\end{table}

\subsubsection{SimCLR}
We report the ResNet-50 pretraining hyperparameters for SimCLR in Table \ref{tab:hypers_simclr}. Linear evaluation is executed for 90 epochs with a learning rate of 0.1, SGD optimizer, batch size of 4096, and no weight decay.

% \usepackage{booktabs}
\begin{table}[h!]
\centering
\begin{tabular}{@{}ll@{}}
\toprule
\textbf{Hyperparameter} & \textbf{Value} \\ \midrule
architecture           & resnet50       \\
proj-hidden-dim      & 2048           \\
out-dim               & 128            \\
use-bn-in-head      & true           \\
batch-size            & 4096           \\
optim                  & LARS           \\
lr                     & 0.3            \\
sqrt-lr               & false          \\
momentum               & 0.9            \\
weight-decay           & 1e-4           \\
epochs                 & 100            \\
warmup-epochs         & 10             \\
zero-init-residual   & true          \\\bottomrule
\end{tabular}
\caption{}
\label{tab:hypers_simclr}
\end{table}

\subsection{Transfer to Other Datasets and Tasks}
For linear evaluation on the transfer datasets, we simply used the same hyperparameters for linear evaluation on ImageNet (DINO and SimSiam respectively). For finetuning DINO ViT-S/16, we used the hyperparameters reported in Table \ref{tab:hypers_transfer_datasets_dino} and for SimSiam ResNet-50 we used the hyperparameters in Table \ref{tab:hypers_transfer_datasets_simsiam}

% \usepackage{booktabs}
\begin{table*}[h!]
\centering
\begin{tabular}{@{}llllll@{}}
\toprule
\textbf{Hyperparameter} & \textbf{CIFAR10} & \textbf{CIFAR100} & \textbf{Flowers102} & \textbf{iNat 21} & \textbf{Food101} \\ \midrule
lr            & 7.5e-6 & 7.5e-6 & 5e-5  & 5e-5  & 5e-5  \\
weight-decay & 0.05   & 0.05   & 0.05  & 0.05  & 0.05  \\
optimizer     & AdamW  & AdamW  & AdamW & AdamW & AdamW \\
epochs        & 300    & 300    & 300   & 100   & 100   \\
batch-size   & 512    & 512    & 512   & 512   & 512   \\ \bottomrule
\end{tabular}
\caption{Finetuning hyperparameters for DINO ViT-S/16.}
\label{tab:hypers_transfer_datasets_dino}
\end{table*}
% \usepackage{booktabs}
\begin{table*}[h!]
\centering
\begin{tabular}{@{}llllll@{}}
\toprule
\textbf{Hyperparameter} & \textbf{CIFAR10} & \textbf{CIFAR100} & \textbf{Flowers102} & \textbf{iNat 21} & \textbf{Food101} \\ \midrule
lr            & 7.5e-6 & 5e-6 & 5e-4  & 7e-5  & 5e-6  \\
weight-decay & 0.05   & 0.05   & 0.05  & 0.05  & 0.05  \\
optimizer     & AdamW  & AdamW  & AdamW & AdamW & AdamW \\
epochs        & 300    & 300    & 300   & 100   & 100   \\
batch-size   & 512    & 512    & 512   & 512   & 512   \\ \bottomrule
\end{tabular}
\caption{Finetuning hyperparameters for SimSiam and ResNet-50.}
\label{tab:hypers_transfer_datasets_simsiam}
\end{table*}

\subsection{Object Detection and Instance Segmentation}
Our experiments utilized Open MMLab's detection library \cite{mmdetection} for object detection and instance segmentation on COCO \cite{lin-eccv14a}. We followed iBOT's default configuration.
% \usepackage{booktabs}
\begin{table}[h!]
\centering
\begin{tabular}{@{}l|r@{}}
\toprule
\multicolumn{2}{c}{\textbf{Obj. Det. \& Inst. Segm. on COCO}} \\
\midrule
\textbf{Hyperparameter} & \textbf{Value} \\
\midrule
epochs      & 12    \\
batch-size  & 32    \\
lr          & 0.02  \\
\bottomrule
\end{tabular}
\caption{Hyperparameters object detection and  instance segmentation on COCO.}
\label{tab:hypers_voc_coco}
\end{table}

\section{Computational Overhead of \MethodAbbr{}}
\label{appendix:time_complexity}
The additional forward passes that \MethodAbbr{} introduces for the selection phase increase the time complexity of the individual baseline methods. Several possible approaches exist to mitigate this overhead, one of which is to alternate between the vanilla and \MethodAbbr{} training step. We measured the overhead factors for different alternating frequencies (i.e., after how many training steps we use the hard views from \MethodAbbr{}; we refer to this as \emph{step}) for SimSiam, DINO, and iBOT which we report in Table \ref{tab:time_factors_slowdown_comparison}. Below, we propose additional ways that may allow using hard views more efficiently.

\begin{table*}[h]
\centering
    \begin{tabular}{l|cccc}
        \toprule
        \textbf{Step} & \multicolumn{1}{c|}{\begin{tabular}[c]{@{}c@{}}\textbf{SimSiam}\\ (RN50)\end{tabular}} & \multicolumn{1}{c|}{\begin{tabular}[c]{@{}c@{}}\textbf{DINO}\\(ViT-S/16)\end{tabular}} & \multicolumn{1}{c|}{\begin{tabular}[c]{@{}c@{}}\textbf{DINO}\\(RN50)\end{tabular}} & \begin{tabular}[c]{@{}c@{}}\textbf{iBOT}\\(ViT-S/16)\end{tabular} \\
        \midrule
        1 (HVP) & x1.64 & x2.21 & x2.01  & x2.13 \\
        2 & x1.38 & x1.61 & x1.56 & x1.57 \\
        3 & x1.32 & x1.43 & x1.42  & x1.38 \\
        4 & x1.29 & x1.34 & x1.35 & x1.29 \\
        \bottomrule
    \end{tabular}
    \caption{Slowdown factors for \MethodAbbr{} and the alternating training method, where \emph{step} refers to the interval at which \MethodAbbr{} is applied during training (i.e., step=1 refers to full \MethodAbbr{} and step=3 indicates that \MethodAbbr{} is approx. used 33\% of the training time). Measurements are averages across 3 seeds.}
    \label{tab:time_factors_slowdown_comparison}
\end{table*}

As an alternative to the alternating training, we also experimented with a 50\% image resolution reduction for the selection phase but observed that the final performance was negatively affected by it or that baseline performance could not be improved. 

The details of hardware and software used for this analysis are: one single compute node with 8 NVIDIA RTX 2080 Ti, AMD EPYC 7502 (32-Core Processor), 512GB RAM, Ubuntu 22.04.3 LTS, PyTorch 2.0.1, CUDA 11.8. For DINO’s 2 global and 8 local views (default), applying \MethodAbbr{} with nviews=2 sampled for each original view results in 4 global and 16 local views. Since considering all combinations would yield over 77k unique comparisons ($\binom{4}{2} \times \binom{16}{8}$), to remain tractable, we limit the number of total comparisons to 64. For training experiments that exceed the limit of 8x RTX 2080 Ti, we apply gradient accumulation.

While technically there can be a memory overhead with \MethodAbbr{}, with the number of sampled views chosen in this paper, the backward pass of the methods that compute gradients only for the selected view pair still consumes more memory than the selection part of \MethodAbbr{} (even for 8 sampled views in SimSiam). Note, that selection and the backward computation are never executed at the same time but sequentially.

We emphasize that further ways exist to optimize \MethodAbbr{}'s efficiency which remain to be explored. For instance:
\begin{itemize}
    \item using embeddings of views from “earlier” layers in the networks or
    \item using 4/8 bit low-precision for the view selection or
    \item using one GPU just for creating embeddings and selecting the hardest views while the remaining GPUs are used for learning or
    \item other approaches to derive manual augmentation policies or 
    \item bypassing forwarding of similar pairs.
\end{itemize}

\section{\MethodName{} Objectives}
\subsection{SimCLR}
\label{appendix:simclr_objective}
In this section, we are going to introduce the application of \MethodAbbr{} with the SimCLR objective. Assume a given set of images $\mathcal{D}$, an image augmentation distribution $\mathcal{T}$, a minibatch of $M$ images $\mathbf{x} =\left\{x_i\right\}_{i=1}^M$ sampled uniformly from $\mathcal{D}$, and two sets of randomly sampled image augmentations $A=\left\{t_i \sim \mathcal{T}\right\}_{i=1}^M$ and $B$ sampled from $\mathcal{T}$. We apply $A$ and $B$ to each image in $\mathbf{x}$ resulting in $\mathbf{x}^A$ and $\mathbf{x}^B$. Both augmented sets of views are subsequently projected into an embedding space with $\mathbf{z}^A=g_{\theta}(f_{\theta}(\mathbf{x}^A))$ and $\mathbf{z}^B=g_{\theta}(f_{\theta}(\mathbf{x}^B))$ where $f_{\theta}$ represents an encoder (or backbone) and $g_{\theta}$ a projector network. Contrastive learning algorithms then minimize the following objective function:

\begin{equation}
\label{eq:cl}
     %$\mathcal{L} (z_i, z_j; \theta) 
    %\mathbb{E}_{x\sim D} \mathcal{L} (z_i, z_j; \theta) 
    \mathcal{L(\mathbf{\mathcal{T}, x; \theta})} = -\log \frac{\exp(\text{sim}(\mathbf{z}_i^A, \mathbf{z}_i^B)/\tau)}{\sum_{i \neq j} \exp(\text{sim}(\mathbf{z}_i^A, \mathbf{z}_j^B)/\tau)}
\end{equation}
where $\tau$ denotes a temperature parameter and \emph{sim} a similarity function that is often chosen as cosine similarity. Intuitively, when optimizing $\theta$, embeddings of two augmented views of the same image are attracted to each other while embeddings of different images are pushed further away from each other.

To further enhance the training process, we introduce a modification to the loss function where instead of having two sets of augmentations \(A\) and \(B\), we now have "N" sets of augmentations, denoted as \(\mathcal{A} = \{A_1, A_2, \ldots, A_N\}\). Each set \(A_i\) is sampled from the image augmentation distribution \(\mathcal{T}\), and applied to each image in \(\mathbf{x}\), resulting in "N" augmented sets of views \(\mathbf{x}^{A_1}, \mathbf{x}^{A_2}, \ldots, \mathbf{x}^{A_N}\). 

Similarly, we obtain $N$ sets of embeddings \(\mathbf{z}^{A_1}, \mathbf{z}^{A_2}, \ldots, \mathbf{z}^{A_N}\) through the encoder and projector networks defined as:

\[
\mathbf{z}^{A_i} = g_{\theta}(f_{\theta}(\mathbf{x}^{A_i})), \quad i = 1, 2, \ldots, N
\]

We then define a new objective function that seeks to find the pair of augmented images that yield the highest loss. The modified loss function is defined as:

\[
\mathcal{L}_{\text{max}}(\mathcal{T}, \mathbf{x}; \theta) = \max_{k, l: k \neq l} \mathcal{L}(\mathcal{T}, \mathbf{x}; \theta)_{kl}
\]

where 

\[
\mathcal{L}(\mathcal{T}, \mathbf{x}; \theta)_{kl} = -\log \frac{\exp(\text{sim}(\mathbf{z}_k^{A_k}, \mathbf{z}_k^{A_l})/\tau)}{\sum_{i \neq j} \exp(\text{sim}(\mathbf{z}_i^{A_k}, \mathbf{z}_j^{A_l})/\tau)}
\]

and \(k, l \in \{1, 2, \ldots, N\}\) and \(i, j \in \{1, 2, \ldots, M\}\). 

For each iteration, we evaluate all possible view pairs and contrast each view against every other example in the mini-batch. Intuitively, the pair that yields the highest loss is selected, which is the pair that at the same time minimizes the numerator and maximizes the denominator in the above equation. In other words, the hardest pair is the one, that has the lowest similarity with another augmented view of itself and the lowest dissimilarity with all other examples.

\end{document}